\documentclass[a4paper,fleqn]{cas-sc}

\usepackage[authoryear,round]{natbib}
\usepackage{booktabs}
\usepackage{threeparttable}
\usepackage{rotating}
\usepackage{multirow}
\usepackage{amssymb} 
\usepackage{tocloft} 
\usepackage{caption} 
\usepackage{hyperref} 
\usepackage{indentfirst}
\usepackage{bbding}
\usepackage{amsthm}
\usepackage{bm}
\usepackage{pifont}

\usepackage{graphicx} 
\usepackage{caption}  
\usepackage{tikz}
\usepackage{amsmath}
\usepackage{algorithm}
\usepackage{algorithmic}
\usepackage{appendix}

\usepackage{tabularx}
\usepackage{amsmath}
\usepackage{multirow}
\usepackage{array} 
\usepackage{svg}

\usepackage{type1cm} 
\newcommand{\tinyscriptsize}{\fontsize{6.5pt}{6.5pt}\selectfont}

\usepackage{colortbl} 
\usepackage{xcolor}   
\usepackage{bbding}

\usepackage{makecell}
\usepackage{enumitem}
\usepackage{diagbox}
\usepackage{subcaption}
\usetikzlibrary{calc}

\def\tsc#1{\csdef{#1}{\textsc{\lowercase{#1}}\xspace}}
\tsc{WGM}
\tsc{QE}
\tsc{EP}
\tsc{PMS}
\tsc{BEC}
\tsc{DE}
\definecolor{mygreen}{RGB}{220,240,255} 
\definecolor{myblue}{RGB}{220,250,220}

\begin{document}
\let\WriteBookmarks\relax
\def\floatpagepagefraction{1}
\def\textpagefraction{.001}

\shorttitle{}


  
\title [mode = title]{A Unified Model for Multi-Task Drone Routing in Post-Disaster Road Assessment}  


\author[1]{Huatian Gong}
\ead{huatian.gong@ntu.edu.sg}

\author[2]{Jiuh-Biing Sheu}
\cormark[1]
\ead{jbsheu@ntu.edu.tw}

\author[3]{Zheng Wang}
\ead{drwz@dlut.edu.cn}

\author[4]{Xiaoguang Yang}
\ead{yangxg@tongji.edu.cn}

\author[1]{Ran Yan}
\cormark[1]
\ead{ran.yan@ntu.edu.sg}

\affiliation[1]{organization={School of Civil and Environmental Engineering, Nanyang Technological University},
country={Singapore}}

\affiliation[2]{organization={Department of Business Administration, National Taiwan University},
country={Taiwan}}

\affiliation[3]{organization={School of Maritime Economics and Management, Dalian Maritime University},
country={China}}

\affiliation[4]{organization={The Key Laboratory of Road and Traffic Engineering of the Ministry of Education, Tongji University},
country={China}}






\begin{abstract}
Post-disaster road assessment (PDRA) is essential for emergency response, enabling rapid evaluation of infrastructure conditions and efficient allocation of resources. Although drones provide a flexible and effective tool for PDRA, routing them in large-scale networks remains challenging. Exact and heuristic optimization methods scale poorly and demand domain expertise, while existing deep reinforcement learning (DRL) approaches adopt a single-task paradigm, requiring separate models for each problem variant and lacking adaptability to evolving operational needs. This study proposes a unified model (UM) for drone routing that simultaneously addresses eight PDRA variants. By training a single neural network across multiple problem configurations, UM captures shared structural knowledge while adapting to variant-specific constraints through a modern transformer encoder-decoder architecture. A lightweight adapter mechanism further enables efficient finetuning to unseen attributes without retraining, enhancing deployment flexibility in dynamic disaster scenarios. Extensive experiments demonstrate that the UM reduces training time and parameters by a factor of eight compared with training separate models, while consistently outperforming single-task DRL methods by 6--14\%, heuristic algorithms by 22--42\%, and commercial solvers by 24--82\% in terms of solution quality (total collected information value). The model achieves rapid solutions (1--10 seconds) across networks of up to 1,000 nodes, with robustness confirmed through sensitivity analyses. Moreover, finetuning experiments show that unseen attributes can be effectively incorporated with minimal cost while retaining high solution quality. Validation on a real-world transportation network further demonstrates the model's practical scalability. The proposed UM advances neural combinatorial optimization for time-critical applications, offering a computationally efficient, high-quality, and adaptable solution for drone-based PDRA. The source code for UM is publicly available at \url{https://github.com/PJ-HTU/UM_PDRA}.
\end{abstract}



\maketitle

\section{Introduction}
\label{Introduction}

Disasters, such as earthquakes and hurricanes, pose severe threats to human lives, infrastructure, and economic stability \citep{avishan2023humanitarian}. In the immediate aftermath of such events, uncertainties regarding transportation infrastructure disruptions emerge, encompassing the extent of road network damage, operability, and feasibility of short-term restoration. Therefore, rapid assessment of road network damage is critical for effective emergency response, as it enables timely delivery of supplies, evacuation planning, and resource allocation \citep{akbari2021online, zhang2023robust}. This process, referred to as post-disaster road assessment (PDRA), is hindered by traditional ground-based methods, which are often impeded by impassable roads, hazardous conditions, and delays, making them impractical for time-sensitive disaster scenarios \citep{zhang2023robust, Gong2025Deep}. Unmanned aerial vehicles (or drones) have emerged as a transformative solution for PDRA, offering high mobility, rapid deployment, and the ability to capture high-resolution imagery in hazardous environments \citep{hong2018range, pang2022adaptive, van2025stochastic}. Equipped with advanced sensors, drones can efficiently assess road damage severity, operability, and repair feasibility, providing critical data to guide relief efforts. Examples include their use in post-typhoon Haiyan \citep{ezequiel2015aerial} and the Nepal earthquake \citep{abuali2025innovative}.

Despite these advantages, solving drone routing for large-scale PDRA remains challenging. Traditional approaches, including exact methods (e.g., branch-and-bound algorithms) \citep{huang2013continuous, zhang2023robust, yin2023robust, morandi2024orienteering} and heuristic methods (e.g., simulated annealing) \citep{oruc2018post, glock2020mission, adsanver2025predictive}, suffer from significant limitations in real-world large-scale PDRA. These methods exhibit solution times that grow dramatically with problem complexity, making them ill-suited to time-critical disaster response where delays can be life-threatening. Moreover, developing effective solving algorithms requires extensive domain knowledge and substantial time investment in algorithm design. To address these two issues, \cite{Gong2025Deep} proposed a deep reinforcement learning (DRL) approach for PDRA, which learns high-quality routing strategies directly from data without relying on domain knowledge, achieving rapid inference (1--10 seconds) and outperforming commercial solvers in solution quality. In recent years, DRL has been used to learn neural network-based heuristics for solving combinatorial optimization problems, receiving growing research attention for its potential to generate high-quality solutions with minimal human effort \citep{vinyals2015pointer, bello2016neural, kool2018attention, kwon2020pomo, luo2023neural, zhou2024mvmoe, liu2024multi, berto2024routefinder}. However, this DRL method operates in a single-task paradigm, requiring a separate model for each specific PDRA variant. When problem characteristics change, such as variations in objective functions or constraints, a new model should be trained from scratch, leading to high computational costs and deployment inefficiencies. Furthermore, the existing method lacks adaptability to unseen future attributes in evolving PDRA scenarios.

Inspired by advancements in large language models (LLMs), where a single model can be applied to diverse tasks with remarkable adaptability and deployment efficiency \citep{achiam2023gpt,touvron2023llama,guo2025deepseek,huang2025orlm,ling2026review}, this study extends the philosophy to neural combinatorial optimization for PDRA. The main contributions are summarized as follows:

\begin{itemize}
\item \textbf{Unified framework for multi-task PDRA.} We develop a unified model (UM) that simultaneously solves eight PDRA variants within a single architecture. Unlike existing traditional optimization and single-task DRL methods, which require separate models for each variant, UM leverages multi-task learning to capture shared structural knowledge while adapting to variant-specific constraints. This design reduces training time and parameter size by a factor of eight compared to training eight separate models, significantly improving computational efficiency and deployment feasibility.
\item \textbf{Lightweight finetuning for unseen attributes.} We introduce an adapter layer mechanism that enables efficient finetuning of the pre-trained UM to incorporate new attributes (e.g., multi-depot settings). This mechanism preserves existing knowledge while adapting to new requirements with minimal computational cost, offering flexibility for rapidly evolving disaster response scenarios.
\item \textbf{State-of-the-art performance and scalability.} Through extensive experiments, we demonstrate that UM outperforms single-task DRL methods (by 6--14\%), heuristic algorithms (by 22--42\%), and commercial solvers (by 24--82\%) in terms of solution quality while maintaining rapid inference (1--10 seconds) for networks with up to 1,000 nodes. Sensitivity analyses confirm the model's robustness under varying operational conditions, and validation on the real-world road network underscores the model’s practical scalability.
\end{itemize}

The remainder of this paper is organized as follows. Section~\ref{Related work} reviews the state-of-the-art in PDRA and related optimization approaches. Section~\ref{Preliminaries} introduces the basic problem formulation. Section~\ref{Variants and motivations} outlines the PDRA variants and the motivation for a unified framework. Section~\ref{Development of the unified model} presents the proposed UM. Section~\ref{Finetuning to unseen attributes} describes the adapter-based finetuning mechanism for new attributes. Section~\ref{Experiments} reports numerical experiments. Finally, Section~\ref{Conclusion} concludes the paper and discusses future research directions.

\section{Related work}
\label{Related work}

PDRA has evolved with technological advancements. Traditional assessment methods, such as field surveys \citep{maya2013rapid} and satellite remote sensing \citep{bravo2019use}, are constrained by slow deployment, labor intensity, and low resolution, which hinder timely decision-making in crisis scenarios. To overcome these limitations, drones have emerged as a transformative solution for PDRA. Drones offer strong mobility, cost efficiency, and high-resolution imaging capabilities, enabling rapid evaluation of road damage (e.g., cracks or landslides) and operability, thus becoming a cornerstone of modern PDRA \citep{hong2018range, zhang2023robust,bogyrbayeva2023deep}.

A critical challenge in PDRA is solving drone routing for large-scale road networks. Existing approaches, including exact algorithms and metaheuristics, face two key limitations: computational inefficiency with increasing problem scale and heavy reliance on domain expertise in model and solution approach design \citep{Gong2025Deep, jiang2025large}. For instance, \citet{zhang2023robust} developed a branch-and-price model validated on 107-node road networks, but its solution time grows exponentially as scale increases. Similarly, \citet{adsanver2025predictive}’s heuristic approach, though effective for 148-node road networks, requires extensive domain knowledge and scales poorly for real-world large-scale PDRA. To address these issues, \cite{Gong2025Deep} introduced a DRL framework for PDRA, leveraging an attention-based encoder-decoder model (AEDM). This method eliminates domain knowledge dependencies by learning routing strategies directly from data, achieving rapid inference (1--10 seconds) for networks with up to 1,000 nodes and outperforming commercial solvers by 16--69\%. 

Like vehicle routing problems with diverse variants \citep{luo2023neural, zhou2024mvmoe, liu2024multi, berto2024routefinder}, PDRA involves multiple routing variants shaped by operational constraints: open routes (no depot return), time windows (temporal assessment constraints), and multi-depot scenarios (distributed launch points). However, existing methods, including exact algorithms, heuristics, and AEDM proposed by \cite{Gong2025Deep}, operate in a single-task paradigm, requiring separate models for each variant. Adapting to new constraints (e.g., switching from closed to open routes) necessitates full model redesign, leading to inefficiencies and limited adaptability to emerging PDRA scenarios. Table~\ref{Table 1} compares these methods, highlighting gaps in unified modeling and adaptability. These approaches lack the ability to handle multiple variants or unseen attributes, underscoring the need for flexible frameworks that can address such limitations.

\begin{table}[!t]
\centering
\caption{Comparison of PDRA Methods}
\label{Table 1}
\resizebox{\textwidth}{!}{
\footnotesize
\renewcommand{\arraystretch}{1.3}
\begin{tabular}{@{}cccccccc@{}}
\toprule
\multirow{2}{*}{\textbf{Authors}} & \multirow{2}{*}{\textbf{Objective}} & \multirow{2}{*}{\textbf{Method}} & \multirow{2}{*}{\textbf{Scale\textsuperscript{a}}} & \multicolumn{2}{c}{\textbf{Efficiency}} & \multirow{2}{*}{\textbf{Uni.}} & \multirow{2}{*}{\textbf{Adpt.}} \\
\cmidrule(lr){5-6}
& & & & \textbf{Time} & \textbf{Expert} & & \\
\midrule
\cite{oruc2018post} & Max. value \& profit & Base route heuristic & 44 & Slow & Yes & & \\
\cite{glock2020mission} & Max. info. in time & Neighborhood search & 625 & Slow & Yes & & \\
\cite{zhang2023robust} & Max. reward & Branch-and-price & 107 & Slow & Yes & & \\
\cite{yin2023robust} & Min. cost & Branch-and-price & 45 & Slow & Yes & & \\
\cite{morandi2024orienteering} & Max. prize & Branch-and-cut & 50 & Slow & Yes & & \\
\cite{adsanver2025predictive} & Max. priority & Variable neighborhood & 148 & Slow & Yes & & \\
\cite{Gong2025Deep} & Max. reward & DRL (AEDM) & 1,000 & Fast & No & & \\
\textbf{This study} & \textbf{Max. reward} & \textbf{Unified DRL} & \textbf{1,000} & \textbf{Fast} & \textbf{No} & \checkmark & \checkmark \\
\bottomrule
\multicolumn{8}{l}{\scriptsize \textsuperscript{a}Number of nodes in the road network.}
\end{tabular}
}
\end{table}

\section{Preliminaries}
\label{Preliminaries}

In this section, we briefly introduce the PDRA using drones. For detailed elaboration, please refer to \cite{Gong2025Deep}. To maintain consistency and facilitate easy reference, we adopt their notation as much as possible. We first present the drone routing problem formulation, followed by the network transformation method and instance generation procedure.

\subsection{Drone routing}

This study addresses the PDRA problem, wherein a fleet of $K$ drones is deployed to assess the condition of road links. The road network is modeled as a graph $G = (N, A)$, where $N$ denotes the set of nodes $i, j \in N$ and $A$ represents the set of links $(i, j) \in A$ connecting these nodes. Let $t_{ij}$ denote the time required for a drone to assess link $(i, j)$, and $c_{ij}$ denote the information value derived from such assessment. The objective is to determine optimal assessment routes for $K$ drones to maximize the total value of information collected across network links.  

Each drone is equipped with high-definition cameras and operates under three critical constraints for PDRA. First, each drone has a battery-imposed flight time limit $Q$. Second, the entire assessment operation must be completed within a maximum allowable time $p_{\text{max}}$. Third, due to operational altitude and visual range limitations, a bidirectional road link can only be assessed if a drone flies directly along it \citep{zhang2023robust}. The PDRA problem involves two distinct time considerations: the computational time to generate assessment routes and the operational time for drones to execute those routes.

A challenge in PDRA formulation is the dual-network structure required for efficient drone operations. As illustrated in \autoref{fig 3.1} (a), the dual-network setup comprises two distinct structures: the original road network $G$ where drones assess individual link damage (shown as solid blue lines), and a fully connected auxiliary network $G' = (N, A')$ where every node is directly connected to all other nodes (shown as dashed red lines). This auxiliary network enables drones to bypass physical road links for faster transit between assessment locations, balancing thorough damage assessment with the time-sensitive requirements of PDRA.

\subsection{Network transformation}
\label{Network transformation}

\begin{figure}
\centering
\includegraphics[width=0.6\textwidth]{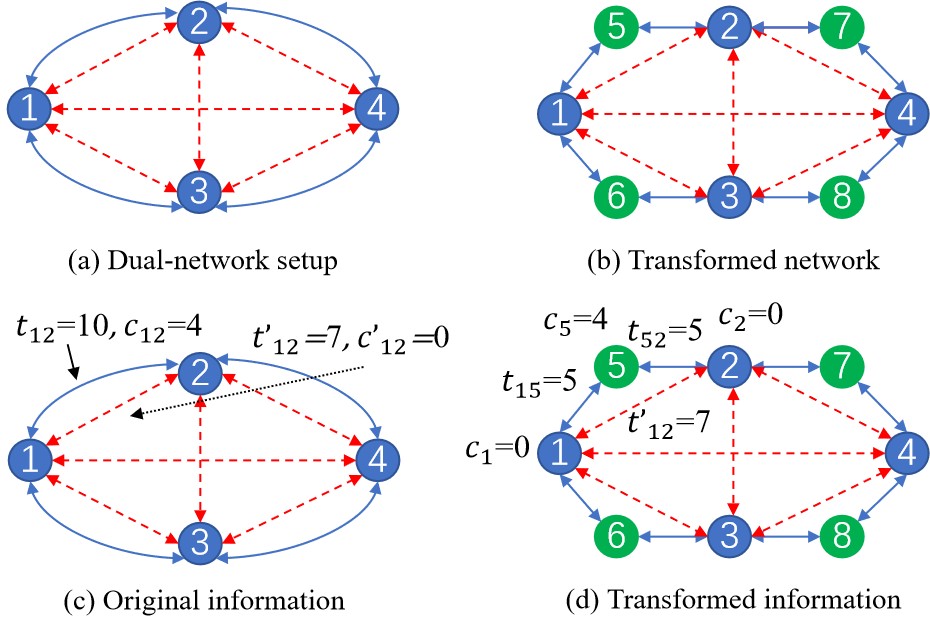}
\caption{Dual-Network Structure and Transformed Node-Based Representation}
\label{fig 3.1}
\end{figure}

To resolve the ambiguities in the dual-network formulation, where both road links and direct routes may connect the same pair of nodes, \cite{Gong2025Deep} developed a network transformation method tailored for PDRA, converting the link-based routing problem into an equivalent node-based formulation. This transformation, illustrated in \autoref{fig 3.1} (b), involves splitting each road link $(i,j) \in A$ by introducing an artificial node $p \in P$, creating two new links: $(i,p)$ and $(p,j)$. The transformation process converts the dual network $G'' = (N, A + A')$ into a reduced network $\bar{G} = ( \bar{N}, \bar{A} ) = (N \cup P, 2A \cup A')$, where $|P| = |A|$. In the transformed network, the ambiguity of multiple edges between nodes is resolved to support accurate PDRA: traversing $(i,p)$ and $(p,j)$ corresponds to assessing the original link $(i,j)$, while direct links in $A'$ remain unchanged for transit purposes. As shown in \autoref{fig 3.1} (c) and (d), the information value of artificial node $p$ is set as $c_p = c_{ij}$, and the traversal time is split as $t_{ip} = t_{pj} = t_{ij}/2$, effectively converting the link-based PDRA routing problem into a node-based problem.

This transformation might allow the PDRA problem to be formulated as a variant of the orienteering problem (OP) \citep{kobeaga2018efficient} where artificial nodes in $P$ carry information values (prizes) and must be visited to collect damage assessment data, while original nodes in $N$ serve as waypoints for routing. However, critical differences exist from standard OP to accommodate PDRA needs: (1) original nodes in $N$ carry no information value, (2) artificial nodes in $P$ are isolated from each other and connect only to specific nodes in $N$, and (3) artificial nodes in $P$ are exclusive (each can be visited by only one drone) to avoid redundant assessments, whereas nodes in $N$ can be revisited, enabling loop formations in drone paths to enhance PDRA efficiency. This characteristic differs from OP's Miller-Tucker-Zemlin (MTZ) formulation \citep{miller1960integer}, which eliminates loops to maintain route connectivity. These architectural distinctions indicate that existing OP methods (whether exact or approximate) cannot be readily adapted. This underscores a prevalent challenge with conventional algorithmic approaches. Even modest alterations to problem characteristics frequently necessitate extensive reformulation, demanding considerable effort and deep domain expertise \citep{ luo2023neural, zhou2024mvmoe, liu2024multi, berto2024routefinder,yang2025heuragenix,liu2025eoh}.

\subsection{Instance generation}

Because large-scale real-world road network datasets suitable for training deep learning models are scarce, \cite{Gong2025Deep} developed a synthetic road network generation method specifically tailored for drone routing problems in PDRA. The generation process follows a four-stage pipeline: (1) \textit{Grid Network Initialization}, which creates a uniform square grid of $ N $ nodes over the coordinate space $[0,1]^2$; (2) \textit{Link Pruning}, which randomly removes edge subsets while maintaining network connectivity to simulate real-world road networks; (3) \textit{Node Perturbation}, which introduces bounded random perturbations to node coordinates to avoid regularity, ensuring the synthetic networks better reflect the irregularities of actual road networks; and (4) \textit{Attribute Assignment}, which computes Euclidean lengths for remaining edges and assigns random damage information values to transformed artificial nodes.

\section{Variants and motivations}
\label{Variants and motivations}
  
Drone routing for PDRA encompasses diverse problem variants, each reflecting unique operational constraints or environmental realities in post-disaster scenarios. These variants as listed below extend the preliminary PDRA \citep{Gong2025Deep} and shape the need for a UM to enhance practical applicability.  

\begin{enumerate}[label=\alph*)]
\item \textbf{Open Routes (OR)}.  
In standard PDRA formulations, drones return to a depot after completing routes (closed routes). However, post-disaster scenarios may require open routes where drones may land at temporary sites instead of returning to the original depot. This configuration prioritizes the rapid collection of road condition data over mandatory depot return, thereby optimizing the operational efficiency of PDRA missions.

\item \textbf{Time Windows (TW)}. 
Road links in PDRA may have time windows with latest allowable assessment times $l_{ij}$ (where $l_{ij}$ denotes the late time window for link $(i,j)$) reflecting when assessment must be completed to remain valuable. For instance, operational priorities in PDRA might require critical links (e.g., hospital access roads) to be assessed within the first assessment phase. Drones would need to traverse link $(i,j)$ before time $l_{ij}$ to collect valid PDRA data, adding temporal feasibility checks that could complicate path sequencing.

\item \textbf{Multi-Depots (MD)}.
Large disaster zones in PDRA may have multiple depots (e.g., fire stations or emergency camps) where drones can launch. Formally, let $D \subseteq N$ be a set of depots ($|D| \geq 2$); drones might start at a depot $d \in D$ and may return to the depot (or none, for the open route) to streamline PDRA operations across vast areas.
\end{enumerate}

The above variants create distinct problem structures for PDRA. The presence or absence of the three attributes results in 8 variants as shown in \autoref{Table 2}. Existing methods, including exact methods, heuristics, and AEDM proposed by \cite{Gong2025Deep}, operate in a single-task paradigm, requiring separate models for each PDRA variant, which hinders practical deployment. A UM capable of solving all PDRA variants within a single model addresses these shortcomings by reducing computational overhead and improving deployment flexibility. Moreover, operational environments may introduce future, previously unseen attributes in evolving PDRA scenarios. In this study, MD is deliberately regarded as such a future attribute for PDRA. After training UM (discussed in Section \ref{Development of the unified model}), the attribute can be incorporated via a lightweight adapter layer mechanism (discussed in Section \ref{Finetuning to unseen attributes}), enabling efficient finetuning without full retraining. This design ensures adaptability to evolving post-disaster response requirements, a critical feature for robust PDRA systems.

\begin{table}[!]
\centering
\caption{PDRA Problem Variants by Attribute Combinations}
\label{Table 2}
\begin{tabular}{ccccc}
\toprule
Variant & \begin{tabular}[c]{@{}c@{}}Open Route\\ (OR)\end{tabular} & \begin{tabular}[c]{@{}c@{}}Time Window\\ (TW)\end{tabular} & \begin{tabular}[c]{@{}c@{}}Multi-depots\\ (MD)\end{tabular} \\ 
\midrule
PDRA-Basic & & & \\
PDRA-OR & \checkmark & & \\
PDRA-TW & & \checkmark & \\
PDRA-OR-TW & \checkmark & \checkmark & \\ 
PDRA-MD & & & \checkmark \\ 
PDRA-OR-MD & \checkmark & & \checkmark \\ 
PDRA-TW-MD & & \checkmark & \checkmark \\ 
PDRA-OR-TW-MD & \checkmark & \checkmark & \checkmark \\ 
\bottomrule
\end{tabular}
\end{table}

\section{Development of the unified model}
\label{Development of the unified model}

\subsection{Model architecture}
\label{Model architecture}

\begin{figure}
\centering
\includegraphics[width=0.95\textwidth]{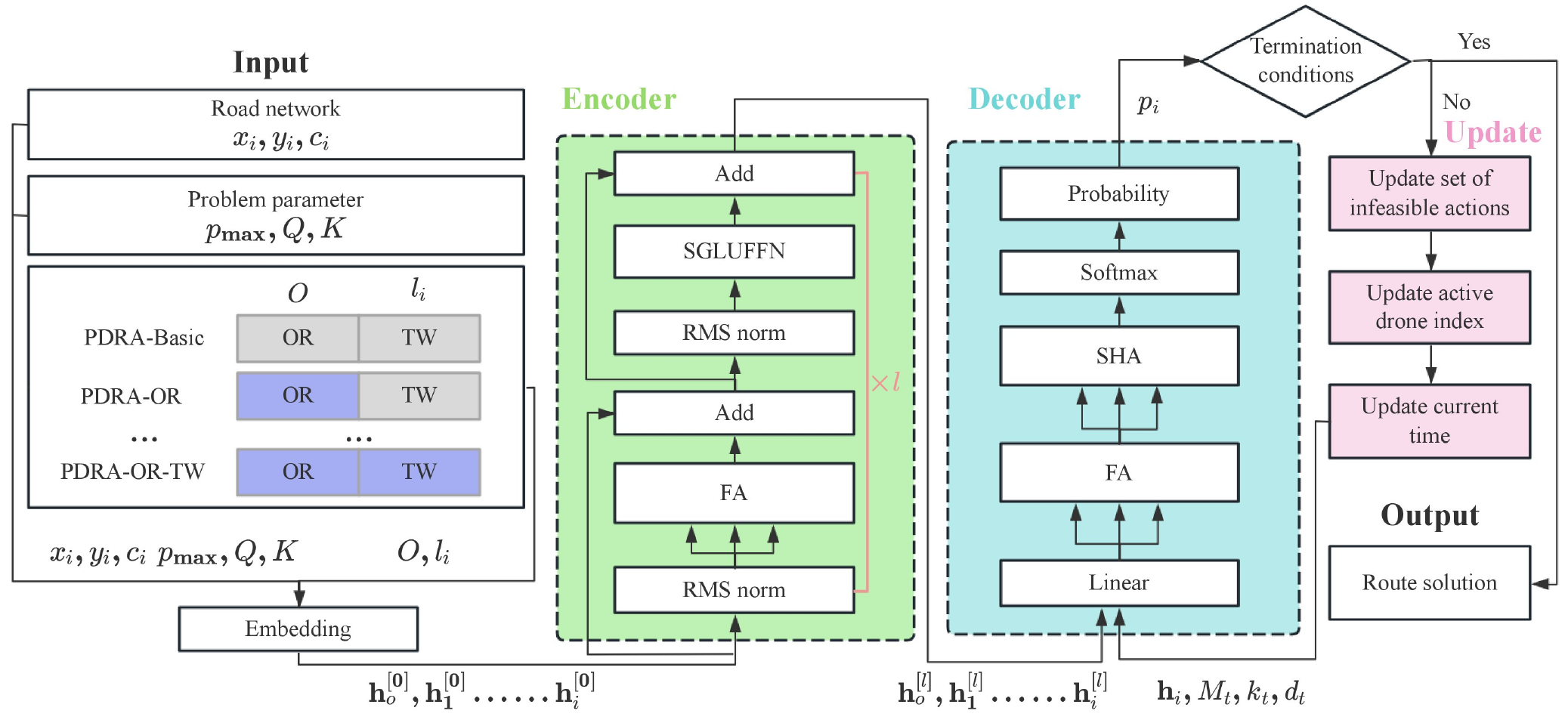}
\caption{Architecture of the Unified Model for Multi-Task PDRA. The unified model processes road network data, problem parameters, and variant attributes through five main components: (1) Input layer embeds node features and problem configurations, distinguishing road network nodes and depot nodes; (2) Encoder transforms embeddings into high-level contextual representations using modern transformer layers with RMS normalization, FlashAttention, and SGLUFFN; (3) Decoder constructs solutions autoregressively through probability computation, softmax selection, and single-head attention mechanisms; (4) Update module maintains solution feasibility by managing infeasible action masks, active drone indices, and current time states; (5) Output produces feasible multi-drone routes satisfying variant-specific constraints with appropriate termination conditions.}
\label{fig 4.1}
\end{figure}

The overall model architecture is illustrated in Figure~\ref{fig 4.1}, which depicts a unified framework capable of handling multiple PDRA variants through a modern encoder-decoder structure. The architecture comprises five main components: input, encoder, decoder, update and output. Each component is detailed in the following subsections.

\subsubsection{Input}
\label{Input}

The input to the model consists of three main parts: road network data, problem parameters, and problem attributes. For any real-world road network, each node $ i \in N $ is linked with geographic coordinates $(x_i, y_i)$ and the corresponding lengths of the connected roads. Through the network transformation method outlined in Section~\ref{Network transformation}, the dual-network structure is converted into an equivalent node-based network. In this transformed network, every node $i \in \bar{N}$ has clearly defined coordinates $(x_i, y_i)$ and information values $c_i$. Problem parameters include operational constraints: maximum assessment time $p_{\max}$, battery flight time limit $Q$, and number of available drones $K$. The problem attributes cover OR and TW, while MD is treated as an unseen attribute (discussed in Section \ref{Finetuning to unseen attributes}). During training, the OR attribute is represented by a binary scalar $O$ ($O = 0$ indicates an open route, and $O = 1$ indicates a closed route). For the TW attribute, since the link-based problem is transformed into a node-based equivalent, the latest time windows $l_{ij}$ specific to each link are also transformed into node-specific values $l_i$. To enhance the model's robustness, the OR and TW attributes are activated independently in each training batch: the value of $O$ is set to 0 or 1 with an equal probability of 50\%; and the time windows are either set to their actual values or to $l_i = \infty$ (meaning that this TW attribute is not considered) with an equal probability of 50\%. An embedding layer is used to map raw node features into a latent space, and the nodes are categorized into two types:

\begin{enumerate}[label=\alph*)]
    \item \textbf{Road network nodes} ($\bar{N}$): these include the original nodes ($N$) and artificial nodes ($P$). Each node $ i \in \bar{N} $ is defined by its coordinates $(x_i, y_i)$ and information value $ c_i $, where $ c_i = 0 $ for nodes $ i \in N $ and $ c_i > 0 $ for artificial nodes $ i \in P $. The initial embeddings are calculated as follows:
    \begin{equation}
        \mathbf{h}_i^{[0]} = [x_i, y_i, c_i, l_i]\,\mathbf{W} + \mathbf{b}, \quad \forall i \in \bar{N},
    \end{equation}
    where $[\cdot]$ stands for feature concatenation, $\mathbf{W} \in \mathbb{R}^{4 \times d}$ and $\mathbf{b} \in \mathbb{R}^d$ are learnable parameters, and $\mathbf{h}_i^{[0]} \in \mathbb{R}^d$ represents the initial embedding of node $ i $.
    
    \item \textbf{Depot node} ($o$): this node acts as both the starting point and the end point, and it combines its coordinates $(x_o, y_o)$ with global parameters $(p_{\max}, Q, K, O)$:
    \begin{equation}
    \mathbf{h}_o^{[0]} = [x_o, y_o, p_{\max}, Q, K, O]\,\mathbf{W}_o + \mathbf{b}_o,
    \end{equation}
    where $\mathbf{W}_o \in \mathbb{R}^{6 \times d}$ and $\mathbf{b}_o \in \mathbb{R}^d$ are learnable parameters specific to the depot node, resulting in the initial embedding $\mathbf{h}_o^{[0]} \in \mathbb{R}^d$.
\end{enumerate}

This heterogeneous embedding approach clearly differentiates between node types and incorporates global operational constraints into the depot node's representation, thereby providing a unified feature space for subsequent processing.

\subsubsection{Encoder}
\label{Encoder}

\begin{figure}
\centering
\includegraphics[width=0.5\textwidth]{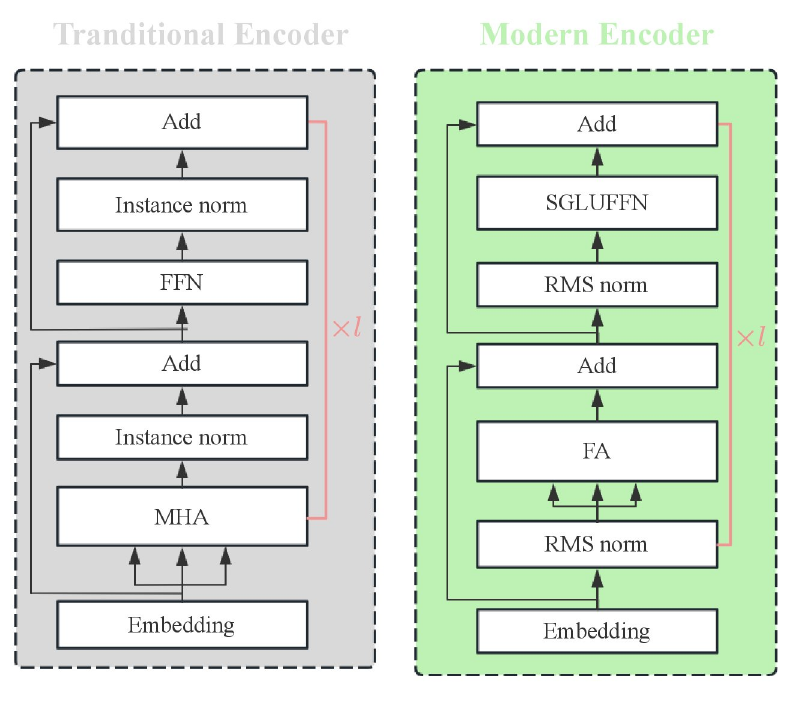}
\caption{Comparison between the Traditional Transformer Encoder Architecture and the Proposed Modern Encoder}
\label{fig 4.2}
\end{figure}

The encoder converts embedded input features into high-level contextual representations through a stack of $l$ transformer layers. In the traditional transformer encoder \citep{waswani2017attention, kool2018attention,kwon2020pomo,zhang2020multi,luo2023neural,zhou2024mvmoe,liu2024multi}, each layer consists of two sublayers: (1) a multi-head attention (MHA) mechanism that models pairwise interactions between nodes, and (2) a position-wise feed-forward network (FFN) that independently refines the representation of each node. Each sublayer is followed by residual connections and instance normalization to ensure stable training and improved generalization. However, traditional transformer architectures face computational inefficiencies when processing large-scale routing problems, particularly in terms of memory usage and training stability. The standard attention mechanism exhibits quadratic complexity with sequence length, while conventional normalization techniques may not provide optimal gradient flow for deep networks. Additionally, traditional ReLU-based FFN have limited expressiveness for capturing complex non-linear relationships in routing optimization. To address these issues, our model adopts a modern transformer encoder \citep{berto2024routefinder}, inspired by advancements in LLM \citep{nguyen2024sequence, dubey2024llama}, with four key enhancements (illustrated in \autoref{fig 4.2}):

\begin{enumerate}[label=\alph*)]
    \item Root Mean Square (RMS) normalization replaces instance normalization, providing better training stability and reducing computational overhead \citep{zhang2019root}.
    \item Pre-normalization configuration shifts the normalization step to before residual connections, enhancing gradient flow and accelerating convergence \citep{jiang2023pre}.
    \item FlashAttention (FA) is used in all MHA layers to improve efficiency when processing large input graphs without losing numerical accuracy \citep{dao2022flashattention, dao2023flashattention}.
    \item Swish Gated Linear Unit Feed-Forward Network (SGLUFFN), an extension of the Gated Linear Unit \citep{dauphin2017language}, replaces the traditional ReLU-based FFN, strengthening the model’s ability to capture complex non-linear relationships \citep{shazeer2020glu}.
\end{enumerate}

Formally, for node $i \in \bar{N}$ and depot node $o$, the computations in the $l$-th encoder layer are as follows:

\begin{equation}
\label{equation 3}
\bar{\mathbf{h}}_i = \text{RMS}\!\left( \mathbf{h}_i^{[l]} \right),
\end{equation}

\begin{equation}
\label{equation 4}
\hat{\mathbf{h}}_i = \mathbf{h}_i^{[l]} + \text{FA}\!\left( \bar{\mathbf{h}}_o, \bar{\mathbf{h}}_1, \dots, \bar{\mathbf{h}}_i \right),
\end{equation}

\begin{equation}
\label{equation 5}
\mathbf{h}_i^{[l+1]} = \hat{\mathbf{h}}_i + \text{SGLUFFN}\!\left( \text{RMS}(\hat{\mathbf{h}}_i) \right),
\end{equation}
where $\mathbf{h}^{[l]} \in \mathbb{R}^d$ denotes the input vector to the $l$-th layer. Detailed formulations of RMS, FA and SGLUFFN are in Appendix \ref{A. Model component details}, and ablation studies in Appendix Equation~\eqref{B. Ablation study} validate the superiority of this modern architecture over traditional transformers.

\subsubsection{Decoder}
\label{Decoder}

The decoder's main role is to build solutions step by step by fusing the final output embeddings $\mathbf{h}_i^{[l]}$ from the encoder with the current state representation at each time step $t$. A context embedding mechanism adjusts the query space dynamically by merging the embedding of the current problem node with relevant state details. This context embedding $\mathbf{h}_c^{(t)}$ is obtained through a linear projection of combined features: the node embedding from the encoder's last layer ($\mathbf{h}_i^{[l]}$) and dynamic state features like the current time $d_t$ and the index $k_t$ of the active drone. In mathematical terms:
\begin{equation}
\label{equation 6}
\mathbf{h}_c^{(t)} = 
\begin{cases} 
\left[ \mathbf{h}_{i}, d_t, k_t \right] \mathbf{W}_c + \mathbf{b}_c & t \geq 1, \\
\left[ \mathbf{h}_{o}, d_t, k_t \right] \mathbf{W}_c + \mathbf{b}_c & t = 0.
\end{cases}
\end{equation}
Here, $\mathbf{h}_{i}$ is the final-layer encoder embedding of the node chosen at the previous time step $i$; $\mathbf{h}_{o}$ is the depot embedding from the encoder's final layer; $d_t$ and $k_t$ respectively capture the temporal state and the index of the active drone; and $\mathbf{W}_c \in \mathbb{R}^{(d+2) \times d}$ and $\mathbf{b}_c \in \mathbb{R}^d$ are learnable projection parameters. For clarity, the encoder layer index is omitted, and all variables from the encoder refer to the output of the final encoder layer $[l]$. This context embedding setup helps the decoder keep track of both the current solution state and operational constraints when generating subsequent actions. By explicitly including the temporal progression ($d_t$) and drone index ($k_t$), the model can adapt its decision policy according to the changing characteristics of the solution across different PDRA variants. Subsequently, the FA layer and a single-head attention (SHA) layer are utilized, expressed as:
\begin{equation}
\label{equation 7}
{\bf{\bar h}}_c^{(t)} = {\rm{FA}} \left({{\bf{h}}_c^{(t)}, \mathbf{h}_{o}, \bf{h}}_1,{\bf{h}}_2, \cdots ,{\bf{h}}_i, M_{t} \right),
\end{equation}
\begin{equation}
\label{equation 8}
\mathbf{u} = u_o,u_1, \cdots,u_{i} = {\rm{SHA}} \left({{\bf{\bar h}}_c^{(t)}, \mathbf{h}_{o}, \mathbf{h}}_1,{\mathbf{h}}_2, \cdots ,{\mathbf{h}}_i, M_{t} \right),
\end{equation}
where $M_{t}$ represents the set of infeasible actions at the current time step $t$, which will be elaborated on later. The probability of selecting node $i$ at time step $t$ is calculated using the softmax function:
\begin{equation}
\label{equation 9}
p_i = \text{Softmax}(C \cdot \text{tanh}(\bf{u})),
\end{equation}
where $C$ acts as a clipping parameter for the tanh function, boosting solution exploration in the search space \citep{bello2016neural, berto2024routefinder}. Once the probability distribution $p_i$ is obtained, the current time step $t$ is concluded, and the state is updated to $t+1$, enabling the decoder to build solutions in an autoregressive manner that adapts to diverse PDRA variants. Detailed mathematical formulations of SHA and Softmax are also provided in Appendix \ref{A. Model component details}.

At each decoding step $t$, the decoder generates a probability distribution $p_i$ over the set of feasible nodes, representing the likelihood of selecting node $i$ as the next visiting location. Sampling or greedy selection based on $p_i$ sequentially constructs a drone’s route. Formally, the decoder outputs a solution $\Pi = (\pi^1,\pi^2,\dots,\pi^K)$, where $\pi^k = (\pi^k_1, \pi^k_2, \dots, \pi^k_{t})$ denotes the ordered sequence of nodes visited by drone $k$. Each route starts from the depot, visits a subset of nodes, and terminates according to the variant-specific constraints and route type (closed or open routes for OR attribute).

\subsubsection{Update}

The update mechanism maintains solution feasibility and tracks the routing state across decoding steps. Two key components are updated at each time step $t$: (1) the set of infeasible actions $M_t$ to ensure constraint compliance, and (2) the system state variables including the active drone index $k_t$ and current time $d_t$. These updates enable the decoder to generate valid solutions across all PDRA variants while adapting to their specific operational constraints.

\paragraph{Infeasible actions for $M_t$.} The decoder employs a masking mechanism that sets $u_i = -\infty$ for infeasible actions at time step $t$. The set of infeasible actions $M_t$ includes the following categories, tailored to handle the attributes of PDRA variants (OR and TW):

\begin{enumerate}[label=\alph*)]
\item Network connectivity constraint: The next node is not connected to the current node at time step $t$ in the transformed road network. This masking ensures that the decoder's output adheres to the routing rules of the road network.

\item Information collection constraint: Nodes with damage information values (i.e., belonging to the set $P$) that have been visited at the current time step $t$ are masked. This masking prevents redundant collection of damage information values from links.

\item Time window constraint: For problems with the TW attribute, nodes are masked if they cannot be visited within their specified time windows. Specifically, node $i$ is masked if $d_t + t_{ji} > l_i$, where $d_t$ is the current time, $t_{ji}$ is the travel time from the previous node $j$ to node $i$, and $l_i$ represents the latest allowable time window for node $i$.

\item Flight time and battery constraints: Each drone must complete its mission within the specified assessment time limit $p_{\max}$ and battery flight time limit $Q$. The masking rules are determined by both the route type (OR attribute) and node characteristics.
   \begin{enumerate}[label=d\arabic*)]
   \item For closed routes, when the next candidate node $i \in N$ (network nodes without damage information) is connected to the depot, node $i$ is masked if $d_t + t_{ji} + t_{io} > \min(p_{\max}, Q)$, where $t_{io}$ denotes the return flight time from node $i$ to depot $o$. When the candidate node $i \in P$ (artificial nodes with damage information) is not directly connected to the depot, it is masked if $d_t + 2t_{ji} + t_{io} > \min(p_{\max}, Q)$, where the coefficient 2 accounts for the traversal time splitting described in Section~\ref{Network transformation}.
   \item For open routes, the masking conditions simplify as depot return is not required. Specifically, nodes $i \in N$ are masked when $d_t + t_{ji} > \min(p_{\max}, Q)$, while nodes $i \in P$ are masked when $d_t + 2t_{ji} > \min(p_{\max}, Q)$.
   \end{enumerate}
\end{enumerate}

These masking mechanisms collectively ensure adherence to the transformed network's structural constraints, prevent redundant information collection, enforce temporal feasibility, and maintain operational time limits across both open and closed route configurations, thereby guaranteeing feasible drone routing solutions for all PDRA variants.

\paragraph{Active drone index for $k_t$.} After each node selection, the decoder updates the system state to maintain consistency across multi-drone operations, accommodating varying fleet sizes without requiring variant-specific adjustments. The update rule for the active drone index $k_{t}$ is designed to handle both closed and open route configurations:
\begin{equation}
k_{t+1} = 
\begin{cases} 
k_{t} + 1 & \text{if route termination condition is met} \\
k_{t} & \text{otherwise}
\end{cases},
\end{equation}
where route termination is defined as the selected node being the depot $o$ for the closed route, or the drone having completed its maximum allowable mission time or visited all accessible nodes for the open route.

\paragraph{Current time for $d_t$.} The current time $d_{t}$ is updated using the same route termination condition as defined above, ensuring temporal consistency across all variants:
\begin{equation}
d_{t+1} = 
\begin{cases} 
0 & \text{if route termination condition is met} \\
d_t + t_{ji} & \text{otherwise}
\end{cases}.
\end{equation}

\subsubsection{Output}

The decoder produces a feasible solution $\Pi = (\pi^1, \pi^2, \ldots, \pi^K)$ and terminates based on route-type-specific criteria. For closed routes, termination occurs when either all $K$ drones have completed their routes by returning to the depot, or all assessable nodes in $P$ have been visited and the active drone returns to the depot. For open routes, termination occurs when either all drones have been deployed and exhausted their available time, or all nodes in $P$ have been assessed. These variant-adaptive termination conditions ensure solution validity across all PDRA configurations while maintaining consistency with the masking constraints defined in the update phase.

\subsection{Multi-task training}  
\label{Multi-task training}

UM employs policy optimization with multiple optima (POMO) adapted for multi-task learning to simultaneously handle all PDRA variants within a single model \citep{kwon2020pomo,liu2024multi,zhou2024mvmoe,berto2024routefinder}. This approach eliminates the computational overhead of training separate models for each variant while leveraging shared knowledge across problem structures.

We implement a batch-wise strategy where each training batch focuses on consistent attribute configurations, ensuring uniform decoder sequence lengths within batches to facilitate efficient parallel processing and stable gradient computation \citep{liu2024multi,zhou2024mvmoe}. Attributes are stochastically activated with 50\% probability to enhance model robustness across diverse constraint configurations, allowing the model to learn adaptive strategies for both OR and TW scenarios. This stochastic activation balances training exposure and prevents bias toward any specific attribute. A critical challenge in multi-task learning is reward discrepancy across attribute combinations; for example, scenarios with TW may yield different reward scales compared to those without. To mitigate training biases, we implement a normalized reward strategy combining exponentially moving average (EMA) with Z-score normalization, following the approach established in \cite{Gong2025Deep}. This normalization anchors rewards to their historical trajectories within each attribute regime, enabling stable multi-task convergence across diverse PDRA variants. Algorithm \ref{algorithm} summarizes the complete multi-task training framework. The unified architecture learns shared representations across PDRA variants while adapting to variant-specific constraints through the attention mechanism and masking strategies, achieving superior performance compared to specialized single-task models.

\begin{algorithm}[ht]
\caption{Multi-Task POMO Training}
\label{algorithm}
\begin{algorithmic}[1]
\STATE \textbf{Input:} Epochs $E$, iterations $T$, batch size $B$, parameter ranges $\{(p_{\text{max}}, Q, K)\}$
\STATE \textbf{Initialize:} Policy network $\theta$
\FOR{$e = 1$ to $E$}
    \FOR{$t = 1$ to $T$}
        \STATE Generate $B$ problem instances
        \STATE Sample a parameter combination $(p_{\text{max}}, Q, K)$
        \STATE Activate OR, TW attributes with probability 50\%
        \STATE Compute solutions via $\theta$-parameterized policy
        \STATE Normalize rewards using EMA with Z-score normalization
        \STATE Update $\theta$ using Adam optimizer via policy gradients
    \ENDFOR
\ENDFOR
\STATE \textbf{Return:} Optimized parameters $\theta$
\end{algorithmic}
\end{algorithm}

\section{Finetuning to unseen attributes}
\label{Finetuning to unseen attributes}

The dynamic nature of post-disaster environments demands adaptive models capable of integrating previously unseen attributes without necessitating full retraining. To validate the adaptability and practical deployment flexibility of the proposed UM, we propose a lightweight finetuning mechanism that enables efficient adaptation to emerging PDRA requirements. For this study, the MD attribute is selected as a representative unseen attribute, as it introduces fundamental structural modifications to the routing problem while maintaining compatibility with existing PDRA variants. Thus, MD serves as a representative unseen attribute and is excluded from the initial training phase to simulate the realistic scenario where new operational requirements emerge post-deployment.

The finetuning mechanism leverages efficient adapter layers, drawing inspiration from recent advancements in parameter-efficient transfer learning \citep{berto2024routefinder}. The core principle involves parameter augmentation with zero-initialized entries; this design choice ensures the preservation of pre-trained knowledge while accommodating new attribute-specific requirements. For integrating a new attribute, the finetuning mechanism requires only minimal modifications to the input embedding and the decoder's context embedding. Taking MD as an example, these modifications are as follows. The original depot node embedding formulation:
\begin{equation}
\label{equation 12}
\mathbf{h}_o^{[0]} = [x_o, y_o, p_{\max}, Q, K, O]\mathbf{W}_o + \mathbf{b}_o,
\end{equation}
requires extension to handle multiple depots $D \subseteq N$ where $|D| \geq 2$. Rather than modifying the existing learned parameters $\mathbf{W}_o \in \mathbb{R}^{6 \times d}$ and $\mathbf{b}_o \in \mathbb{R}^d$, architectural consistency is maintained by computing each depot's embedding ($d \in D$) using the same projection matrices with depot-specific coordinates $(x_d, y_d)$. This design preserves the semantic understanding of depot-related features acquired during pre-training while enabling the representation of multiple depot instances. To incorporate MD-specific state information for effective multi-depot routing decisions, the decoder's context embedding mechanism is enhanced as follows:

\begin{equation}
\label{equation 13}
\mathbf{h}_c^{(t)} = 
\begin{cases} 
\left[ \mathbf{h}_{i}, d_t, k_t, x_d, y_d \right] \mathbf{W}_c' + \mathbf{b}_c' & t \geq 1, \\
\left[ \mathbf{h}_{o}, d_t, k_t , x_d, y_d \right] \mathbf{W}_c' + \mathbf{b}_c' & t = 0.
\end{cases}
\end{equation}

Compared with the original context embedding in Equation~\eqref{equation 6}, the enhanced context embedding in Equation~\eqref{equation 13} adds two dimensions $(x_d, y_d)$. These two additional dimensions serve specific operational functions, particularly capturing origin depot coordinates to enforce return constraints in closed-route configurations, thereby ensuring spatial awareness for depot-specific return requirements. Correspondingly, the original projection matrices $\mathbf{W}_c \in \mathbb{R}^{(d+2) \times d}$ and $\mathbf{b}_c \in \mathbb{R}^d$ are expanded via zero-initialized parameter expansion:

\begin{equation}
\mathbf{W}_c' = \begin{bmatrix} \mathbf{W}_c \\ \mathbf{0}_{2 \times d} \end{bmatrix}; \quad \mathbf{b}_c' = \begin{bmatrix} \mathbf{b}_c \\ \mathbf{0}_{2 \times 1} \end{bmatrix}.
\end{equation}

The decoder's termination conditions require modification to accommodate multi-depot scenarios. The updated termination criteria are compatible with both OR and MD attributes. For closed routes with MD, termination occurs when all drones have completed their routes and returned to their assigned depots, or when all nodes with damage information values have been visited and the active drone returns to its assigned depot. For open routes with MD, termination occurs when all drones have been deployed and completed their missions, or when all nodes with damage information values have been visited. These modified termination conditions ensure valid solution generation across multi-depot configurations while maintaining compatibility with existing route type constraints. By leveraging the shared knowledge encoded in UM's representations, the finetuning process requires significantly fewer training iterations compared to training from scratch, an efficiency critical for rapid deployment in disaster response scenarios where new operational requirements may emerge with limited preparation time.

\section{Experiments}
\label{Experiments}

This section evaluates the proposed UM through extensive experiments across multiple PDRA variants and network scales, examining solution quality, computational efficiency, scalability, and adaptability to unseen attributes.

\subsection{Experiment setup and benchmarks}
\label{Experimental setup and benchmarks}

The experimental configuration encompasses model architecture specifications, problem parameter settings, training infrastructure, benchmark methods, and evaluation protocol.

\textbf{Model Parameters}. UM employs an embedding size $d = 128$ and features a modern transformer encoder consisting of $l = 6$ layers with 8-head attention and SGLUFFN with a hidden size of 512. These architectural settings are informed by successful applications in neural combinatorial optimization \citep{kool2018attention, kwon2020pomo, luo2023neural, zhou2024mvmoe, liu2024multi, berto2024routefinder,Gong2025Deep}. UM configuration yields approximately 1.3 million trainable parameters. Training spans 200 epochs (batch size = 64) on 100,000 on-the-fly generated instances to ensure robust learning across diverse problem variants. The Adam optimizer employs an initial learning rate of $10^{-4}$, $L_2$ regularization ($10^{-6}$ weight decay), and learning rate decay (×0.1 at epochs 175 and 195, respectively).

\textbf{Problem Parameters}. UM is trained on 100-node instances, comprising 50 nodes in set $P$ (with damage information values, representing 50 assessable road links) and 50 nodes in set $N$ (without damage information, including one depot). To evaluate generalization capability, testing is conducted on larger instances (200, 400, 600, 800, and 1,000 nodes), each with 50\% nodes in $P$ and 50\% in $N$. Damage information values for $P$ are uniformly sampled from [1, 10] and normalized by division by 10. Practically, drones have a 2-hour battery flight time limit $Q$ and a speed of 60 km/h \citep{zhang2023robust}. Given the time-critical nature of disaster response, assessment time limits $p_{\max}$ are set to 30, 45, and 60 minutes (corresponding to 30, 45, and 60 km flight distances, respectively). Since $Q$ is significantly larger than $p_{\max}$, $Q$ is not considered in subsequent experiments, with a focus on $p_{\max}$. Road network coordinates are normalized to $[0,1]$, with link lengths scaled accordingly. For simplified data generation, training uses $p_{\max} = 2, 3, 4$ corresponding to 30, 45, and 60 minutes in real-world distances. For parameter $K$, training uses 2, 3, or 4 drones. For the latest time windows parameter $l_i$ in the TW attribute, the generation method adopted is consistent with the approach proposed for classical vehicle routing problems in the existing literature \citep{zhao2020hybrid, liu2024multi, zhou2024mvmoe, berto2024routefinder}.

\textbf{Training Infrastructure}. Training is conducted on Google Colab (\url{https://colab.research.google.com}) using an NVIDIA A100 GPU. Each epoch requires approximately 7.2 minutes, resulting in a total training duration of around 24 hours for 200 epochs. The source code for UM is publicly available at \url{https://github.com/PJ-HTU/UM_PDRA}.

\textbf{Benchmark Methods}. To extensively evaluate UM's performance, we establish three categories of baseline methods representing different optimization paradigms: exact optimization, heuristic optimization, and single-task deep learning.

\begin{enumerate}[label=\alph*)]

\item Exact Optimization Baseline. The mathematical formulation for each PDRA variant is provided in Appendix~\ref{C. Mathematical formulation}. We solve these formulations using Gurobi, a state-of-the-art commercial optimizer. While the exact method requires no algorithm design expertise, it necessitate variant-specific model formulations due to differing constraint structures across OR, TW, and MD attributes. To reflect varying urgency levels in disaster response scenarios, Gurobi is evaluated under two time limit settings: 60 seconds (Gurobi-1×60) and 1,800 seconds (Gurobi-30×60), with the best solution reported for each setting.

\item Heuristic Optimization Baseline. We develop a family of two-phase heuristics specifically designed for PDRA variants, with detailed algorithmic descriptions provided in Appendix~\ref{Heuristic algorithms}. These heuristics are inspired by classical OP frameworks \citep{kobeaga2018efficient, yang2025heuragenix} but are substantially modified to accommodate the unique structural and feasibility characteristics of PDRA. Each heuristic consists of: (1) a \textit{greedy construction (GC)} phase that builds initial solutions using a composite scoring function that balances information value collection with temporal urgency, adapted to the transformed network structure where artificial nodes represent road links with exclusive visit requirements, and (2) a \textit{local search (LS)} improvement phase that applies three operators (relocate, exchange, and remove-insert) carefully adapted to handle artificial-node triplet structures while respecting variant-specific constraints. Due to fundamental differences in feasibility rules across variants (OR modifies depot return logic, TW enforces temporal constraints, and MD introduces depot assignment requirements), each variant requires its own tailored heuristic implementation. We denote these heuristics as GC-Variant for construction only and GC+LS-Variant for construction with improvement, where Variant $\in$ \{Basic, OR, TW, OR-TW, MD, OR-MD, TW-MD, OR-TW-MD\}, corresponding to all eight PDRA variants. While these heuristics provide strong baselines, their requirement for hand-crafted domain knowledge and variant-specific redesign exemplifies the limitations that motivate UM's unified learning approach.

\item Single-Task Deep Learning Baseline. We compare UM against single-task attention-based encoder-decoder models (AEDM) from \cite{Gong2025Deep}. Following the single-task paradigm, eight independent AEDM models are trained, one for each PDRA variant: AEDM-Basic, AEDM-OR, AEDM-TW, AEDM-OR-TW, AEDM-MD, AEDM-OR-MD, AEDM-TW-MD, and AEDM-OR-TW-MD. Each model undergoes separate training (200 epochs per variant) to learn variant-specific routing strategies. This single-task approach results in high training costs and deployment complexity: 8 separate models with a total of 10.4 million parameters, 1,600 cumulative training epochs, and 192 hours of training time.
\end{enumerate}

As summarized in Table~\ref{table 3}, the three baseline categories demonstrate distinct limitations that motivate our unified framework. Exact methods require variant-specific mathematical formulations for each PDRA configuration, resulting in high deployment complexity and poor computational scalability despite minimal algorithmic expertise requirements. Heuristic methods demand extensive domain knowledge and variant-specific algorithmic redesign, with each of the eight PDRA variants necessitating tailored implementations that complicate practical deployment. Single-task DRL approaches eliminate the need for domain expertise and achieve rapid performance, but require training separate models for each variant, leading to substantial deployment overhead when comprehensive variant coverage is needed. In contrast, UM consolidates all 8 variants within a single 1.3M parameter model trained for 200 epochs in 24 hours, achieving an 8-fold reduction in training cost while enabling streamlined deployment through unified problem representation.

\textbf{Evaluation Protocol}. Since post-disaster response inherently involves assessing road networks at multiple affected locations (e.g., after earthquakes or floods), we evaluate all methods on their ability to handle multiple problem instances efficiently, where both solution quality and computational speed are critical. Results are averaged over 10 randomly generated instances for each problem configuration. For UM and AEDM, we report total computation time across all 10 instances to reflect practical deployment scenarios, while for Gurobi and heuristics we report average time per instance, which provides these baseline methods a favorable computational comparison basis.

\begin{table}[!t]
\centering
\caption{Comparison of Methodological Requirements and Efficiency}
\label{table 3}
\begin{tabular}{ccccc}
\toprule
Method & \makecell{Models/Formulations \\ per Variant} & \makecell{Deployment \\ Complexity} & \makecell{Total Training \\ (for 8 variants)} & \makecell{Domain \\ Expertise} \\
\midrule
Exact optimization baseline & 8 formulations & High & N/A & Low \\
Heuristic optimization baseline & 8 heuristics & High & N/A & High \\
Single-task deep learning baseline & 8 models & High & 10.4M params, 192h & Low \\
UM (Ours) & 1 model & Low & 1.3M params, 24h & Low \\
\bottomrule
\end{tabular}
\end{table}

\subsection{Computational performance}  
\label{Computational performance} 

\begin{table}[!h]
\tinyscriptsize
\caption{Performance Comparison of Methods in 200-Node Network ($p_{\max}=30$ min, Varying Drone Fleet sizes)}
\label{Table 4}
\begin{tabular}{cccccccccc}
\toprule
Type & Method &$^{\#}$Drone& Time (s) & Value & Gap &$^{\#}$Drone& Time (s) & Value & Gap \\
\midrule
\multirow{18}{*}{PDRA-Basic} & \multirow{2}{*}{Gurobi} & \multirow{9}{*}{2} & 60 & 8.18 & 49.41\% & \multirow{9}{*}{3} & 60 & 9.66 & 56.23\% \\
 &  &  & 30×60 & 16.11 & 0.37\% &  & 30×60 & 21.04 & 4.67\% \\
 & GC-Basic &  & 1 & 11.15 & 31.05\% &  & 2 & 14.63 & 33.71\% \\
 & GC+LS-Basic &  & 19 & 13.14 & 18.74\% &  & 37 & 17.58 & 20.35\% \\
 & AEDM-Basic &  & 1 & 16.28 & -0.68\% &  & 1 & 21.98 & 0.41\% \\
 & AEDM-OR &  & 1 & 14.47 & 10.51\% &  & 1 & 19.61 & 11.15\% \\
 & AEDM-TW &  & 1 & 14.18 & 12.31\% &  & 1 & 18.95 & 14.14\% \\
 & AEDM-OR-TW &  & 1 & 14.07 & 12.99\% &  & 1 & 18.12 & 17.90\% \\
& \cellcolor{myblue}UM & \cellcolor{myblue}& \cellcolor{myblue}1 & \cellcolor{myblue}16.17 & \cellcolor{myblue}0.00\% & \cellcolor{myblue}& \cellcolor{myblue}1 & \cellcolor{myblue}22.07 & \cellcolor{myblue}0.00\% \\
& \multirow{2}{*}{Gurobi} & \multirow{9}{*}{4} & 60 & 9.89 & 63.14\% & \multirow{9}{*}{5} & 60 & 11.67 & 62.60\% \\
 &  &  & 30×60 & 20.78 & 22.55\% &  & 30×60 & 23.15 & 25.80\% \\
 & GC-Basic &  & 6 & 16.87 & 37.12\% &  & 10 & 19.48 & 37.56\% \\
 & GC+LS-Basic &  & 108 & 20.14 & 24.93\% &  & 282 & 22.38 & 28.27\% \\
 & AEDM-Basic &  & 2 & 26.59 & 0.89\% &  & 2 & 30.80 & 1.28\% \\
 & AEDM-OR &  & 2 & 23.59 & 12.08\% &  & 2 & 27.51 & 11.83\% \\
 & AEDM-TW &  & 2 & 22.55 & 15.95\% &  & 2 & 24.67 & 20.93\% \\
 & AEDM-OR-TW &  & 2 & 21.75 & 18.93\% &  & 2 & 23.21 & 25.61\% \\
& \cellcolor{myblue}UM & \cellcolor{myblue} & \cellcolor{myblue}2 & \cellcolor{myblue}26.83 & \cellcolor{myblue}0.00\% & \cellcolor{myblue}& \cellcolor{myblue}2 & \cellcolor{myblue}31.20 &\cellcolor{myblue} 0.00\% \\
 \midrule
\multirow{18}{*}{PDRA-OR} & \multirow{2}{*}{Gurobi} & \multirow{9}{*}{2} & 60 & 8.18 & 57.88\% & \multirow{9}{*}{3} & 60 & 9.76 & 64.33\% \\
 &  &  & 30×60 & 16.11 & 17.04\% &  & 30×60 & 21.04 & 23.10\% \\
 & GC-OR &  & 1 & 14.41 & 25.78\% &  & 2 & 16.38 & 40.13\% \\
 & GC+LS-OR &  & 17 & 16.31 & 16.01\% &  & 39 & 20.14 & 26.39\% \\
 & AEDM-Basic &  & 1 & 18.56 & 4.43\% &  & 1 & 25.87 & 5.45\% \\
 & AEDM-OR &  & 1 & 19.35 & 0.36\% &  & 1 & 27.28 & 0.29\% \\
 & AEDM-TW &  & 1 & 17.02 & 12.36\% &  & 1 & 23.04 & 15.79\% \\
 & AEDM-OR-TW &  & 1 & 17.68 & 8.96\% &  & 1 & 24.34 & 11.04\% \\
& \cellcolor{myblue}UM & \cellcolor{myblue} & \cellcolor{myblue}1 & \cellcolor{myblue}19.42 & \cellcolor{myblue}0.00\% & \cellcolor{myblue} & \cellcolor{myblue}1 & \cellcolor{myblue}27.36 & \cellcolor{myblue}0.00\% \\
& \multirow{2}{*}{Gurobi} & \multirow{9}{*}{4} & 60 & 9.89 & 70.70\% & \multirow{9}{*}{5} & 60 & 11.33 & 71.19\% \\
 &  &  & 30×60 & 20.90 & 38.09\% &  & 30×60 & 23.60 & 39.99\% \\
 & GC-OR &  & 5 & 18.91 & 43.98\% &  & 10 & 23.15 & 41.17\% \\
 & GC+LS-OR &  & 101 & 25.66 & 23.99\% &  & 271 & 30.45 & 22.57\% \\
 & AEDM-Basic &  & 2 & 32.14 & 4.80\% &  & 2 & 37.52 & 4.60\% \\
 & AEDM-OR &  & 2 & 33.72 & 0.12\% &  & 2 & 39.32 & 0.03\% \\
 & AEDM-TW &  & 2 & 25.85 & 23.43\% &  & 2 & 28.56 & 27.38\% \\
 & AEDM-OR-TW &  & 2 & 27.54 & 18.42\% &  & 2 & 35.51 & 9.71\% \\
& \cellcolor{myblue}UM & \cellcolor{myblue} & \cellcolor{myblue}2 & \cellcolor{myblue}33.76 & \cellcolor{myblue}0.00\% & \cellcolor{myblue} & \cellcolor{myblue}2 & \cellcolor{myblue}39.33 & \cellcolor{myblue}0.00\% \\
\midrule
\multirow{18}{*}{PDRA-TW} & \multirow{2}{*}{Gurobi} & \multirow{9}{*}{2} & 60 & 4.28 & 70.11\% & \multirow{9}{*}{3} & 60 & 5.83 & 69.99\% \\
 &  &  & 30×60 & 12.10 & 15.50\% &  & 30×60 & 16.57 & 14.72\% \\
 & GC-TW &  & 2 & 10.51 & 26.61\% &  & 5 & 14.48 & 25.48\% \\
 & GC+LS-TW &  & 23 & 13.14 & 8.24\% &  & 51 & 17.41 & 10.40\% \\
 & AEDM-Basic &  & 1 & 13.59 & 5.10\% &  & 1 & 18.77 & 3.40\% \\
 & AEDM-OR &  & 1 & 12.40 & 13.41\% &  & 1 & 16.98 & 12.61\% \\
 & AEDM-TW &  & 1 & 14.40 & -0.56\% &  & 1 & 19.34 & 0.46\% \\
 & AEDM-OR-TW &  & 1 & 14.22 & 0.70\% &  & 1 & 18.47 & 4.94\% \\
& \cellcolor{myblue}UM & \cellcolor{myblue} & \cellcolor{myblue}1 & \cellcolor{myblue}14.32 & \cellcolor{myblue}0.00\% & \cellcolor{myblue} & \cellcolor{myblue}1 & \cellcolor{myblue}19.43 & \cellcolor{myblue}0.00\% \\
 & \multirow{2}{*}{Gurobi} & \multirow{9}{*}{4} & 60 & 7.02 & 70.76\% & \multirow{9}{*}{5} & 60 & 10.22 & 63.43\% \\
 &  &  & 30×60 & 19.41 & 19.16\% &  & 30×60 & 21.86 & 21.76\% \\
 & GC-TW &  & 12 & 17.45 & 27.32\% &  & 21 & 18.56 & 33.57\% \\
 & GC+LS-TW &  & 131 & 20.14 & 16.12\% &  & 384 & 22.74 & 18.61\% \\
 & AEDM-Basic &  & 2 & 22.78 & 5.12\% &  & 2 & 25.99 & 6.98\% \\
 & AEDM-OR &  & 2 & 19.62 & 18.28\% &  & 2 & 22.62 & 19.04\% \\
 & AEDM-TW &  & 2 & 23.98 & 0.12\% &  & 2 & 27.64 & 1.07\% \\
 & AEDM-OR-TW &  & 2 & 22.98 & 4.29\% &  & 2 & 24.56 & 12.10\% \\
& \cellcolor{myblue}UM & \cellcolor{myblue} & \cellcolor{myblue}2 & \cellcolor{myblue}24.01 & \cellcolor{myblue}0.00\% & \cellcolor{myblue} & \cellcolor{myblue}2 & \cellcolor{myblue}27.94 & \cellcolor{myblue}0.00\% \\
\midrule
\multirow{18}{*}{PDRA-OR-TW} & \multirow{2}{*}{Gurobi} & \multirow{9}{*}{2} & 60 & 6.10 & 64.09\% & \multirow{9}{*}{3} & 60 & 7.64 & 67.48\% \\
 &  &  & 30×60 & 14.26 & 16.02\% &  & 30×60 & 16.57 & 29.46\% \\
 & GC-OR-TW &  & 2 & 11.34 & 33.22\% &  & 7 & 12.56 & 46.53\% \\
 & GC+LS-OR-TW &  & 24 & 13.94 & 17.91\% &  & 58 & 16.85 & 28.27\% \\
 & AEDM-Basic &  & 1 & 15.11 & 11.01\% &  & 1 & 20.86 & 11.20\% \\
 & AEDM-OR &  & 1 & 15.74 & 7.30\% &  & 1 & 21.36 & 9.07\% \\
 & AEDM-TW &  & 1 & 15.51 & 8.66\% &  & 1 & 21.20 & 9.75\% \\
 & AEDM-OR-TW &  & 1 & 16.76 & 1.30\% &  & 1 & 23.10 & 1.66\% \\
& \cellcolor{myblue}UM & \cellcolor{myblue} & \cellcolor{myblue}1 & \cellcolor{myblue}16.98 & \cellcolor{myblue}0.00\% & \cellcolor{myblue} & \cellcolor{myblue}1 & \cellcolor{myblue}23.49 & \cellcolor{myblue}0.00\% \\
 & \multirow{2}{*}{Gurobi} & \multirow{9}{*}{4} & 60 & 9.68 & 66.52\% & \multirow{9}{*}{5} & 60 & 13.09 & 61.17\% \\
 &  &  & 30×60 & 19.41 & 32.88\% &  & 30×60 & 21.86 & 35.13\% \\
 & GC-OR-TW &  & 14 & 18.12 & 37.34\% &  & 26 & 18.44 & 45.28\% \\
 & GC+LS-OR-TW &  & 138 & 20.85 & 27.91\% &  & 410 & 28.14 & 16.50\% \\
 & AEDM-Basic &  & 2 & 26.02 & 10.03\% &  & 2 & 29.92 & 11.22\% \\
 & AEDM-OR &  & 2 & 26.38 & 8.78\% &  & 2 & 30.15 & 10.53\% \\
 & AEDM-TW &  & 2 & 25.94 & 10.30\% &  & 2 & 29.79 & 11.60\% \\
 & AEDM-OR-TW &  & 2 & 28.57 & 1.21\% &  & 2 & 33.45 & 0.74\% \\
& \cellcolor{myblue}UM & \cellcolor{myblue} & \cellcolor{myblue}2 & \cellcolor{myblue}28.92 & \cellcolor{myblue}0.00\% & \cellcolor{myblue} & \cellcolor{myblue}2 & \cellcolor{myblue}33.70 & \cellcolor{myblue}0.00\% \\
 \bottomrule
\end{tabular}
\end{table}

\begin{table}[!]
\tinyscriptsize
\caption{Performance Comparison of Methods in 400-Node Network ($p_{\max}=45$ min, Varying Drone Fleet sizes)}
\label{Table 5}
\begin{tabular}{cccccccccc}
\toprule
Type & Method &$^{\#}$Drone& Time (s) & Value & Gap &$^{\#}$Drone& Time (s) & Value & Gap \\
\midrule
\multirow{18}{*}{PDRA-Basic} & \multirow{2}{*}{Gurobi} & \multirow{9}{*}{2} & 60 & \ding{55} & 100\% & \multirow{9}{*}{3} & 60 & \ding{55} & 100\% \\
 &  &  & 30×60 & 33.98 & 9.00\% &  & 30×60 & 46.89 & 9.27\% \\
 & GC-Basic &  & 3 & 24.17 & 35.27\% &  & 6 & 34.93 & 32.41\% \\
 & GC+LS-Basic &  & 63 & 31.41 & 15.88\% &  & 141 & 40.32 & 21.99\% \\
 & AEDM-Basic &  & 1 & 37.63 & -0.78\% &  & 1 & 51.34 & 0.66\% \\
 & AEDM-OR &  & 1 & 35.77 & 4.20\% &  & 1 & 46.54 & 9.95\% \\
 & AEDM-TW &  & 1 & 31.87 & 14.65\% &  & 1 & 38.31 & 25.87\% \\
 & AEDM-OR-TW &  & 1 & 31.25 & 16.31\% &  & 1 & 37.15 & 28.12\% \\
& \cellcolor{myblue}UM & \cellcolor{myblue} & \cellcolor{myblue}1 & \cellcolor{myblue}37.34 & \cellcolor{myblue}0.00\% & \cellcolor{myblue} & \cellcolor{myblue}1 & \cellcolor{myblue}51.68 & \cellcolor{myblue}0.00\% \\
& \multirow{2}{*}{Gurobi} & \multirow{9}{*}{4} & 60 & \ding{55} & 100\% & \multirow{9}{*}{5} & 60 & \ding{55} & 100\% \\
 &  &  & 30×60 & 50.99 & 20.00\% &  & 30×60 & 49.93 & 31.38\% \\
 & GC-Basic &  & 15 & 40.21 & 36.90\% &  & 30 & 42.60 & 41.46\% \\
 & GC+LS-Basic &  & 350 & 51.21 & 19.65\% &  & 856 & 53.12 & 27.00\% \\
 & AEDM-Basic &  & 2 & 64.16 & -0.66\% &  & 2 & 72.42 & 0.47\% \\
 & AEDM-OR &  & 2 & 58.21 & 8.68\% &  & 2 & 65.28 & 10.28\% \\
 & AEDM-TW &  & 2 & 44.35 & 30.42\% &  & 2 & 47.23 & 35.09\% \\
 & AEDM-OR-TW &  & 2 & 42.47 & 33.37\% &  & 2 & 46.43 & 36.19\% \\
& \cellcolor{myblue}UM & \cellcolor{myblue} & \cellcolor{myblue}2 & \cellcolor{myblue}63.74 & \cellcolor{myblue}0.00\% & \cellcolor{myblue} & \cellcolor{myblue}2 & \cellcolor{myblue}72.76 & \cellcolor{myblue}0.00\% \\
\midrule
\multirow{18}{*}{PDRA-OR} & \multirow{2}{*}{Gurobi} & \multirow{9}{*}{2} & 60 & \ding{55} & 100\% & \multirow{9}{*}{3} & 60 & \ding{55} & 100\% \\
 &  &  & 30×60 & 34.29 & 17.49\% &  & 30×60 & 47.84 & 16.96\% \\
 & GC-OR &  & 4 & 28.07 & 32.46\% &  & 7 & 37.30 & 35.25\% \\
 & GC+LS-OR &  & 68 & 36.14 & 13.04\% &  & 145 & 44.11 & 23.43\% \\
 & AEDM-Basic &  & 1 & 39.61 & 4.69\% &  & 1 & 45.81 & 20.48\% \\
 & AEDM-OR &  & 1 & 41.28 & 0.67\% &  & 1 & 57.57 & 0.07\% \\
 & AEDM-TW &  & 1 & 33.81 & 18.65\% &  & 1 & 40.46 & 29.77\% \\
 & AEDM-OR-TW &  & 1 & 35.48 & 14.63\% &  & 1 & 50.54 & 12.27\% \\
& \cellcolor{myblue}UM & \cellcolor{myblue} & \cellcolor{myblue}1 & \cellcolor{myblue}41.56 & \cellcolor{myblue}0.00\% & \cellcolor{myblue} & \cellcolor{myblue}1 & \cellcolor{myblue}57.61 & \cellcolor{myblue}0.00\% \\
& \multirow{2}{*}{Gurobi} & \multirow{9}{*}{4} & 60 &\ding{55}& 100\% & \multirow{9}{*}{5} & 60 &\ding{55}& 100\% \\
 &  &  & 30×60 & 52.97 & 27.21\% &  & 30×60 & 37.17 & 54.87\% \\
 & GC-OR &  & 17 & 46.40 & 36.23\% &  & 28 & 31.42 & 61.85\% \\
 & GC+LS-OR &  & 361 & 55.14 & 24.23\% &  & 841 & 60.69 & 26.31\% \\
 & AEDM-Basic &  & 2 & 60.97 & 16.22\% &  & 2 & 74.83 & 9.14\% \\
 & AEDM-OR &  & 2 & 72.55 & 0.30\% &  & 2 & 82.27 & 0.11\% \\
 & AEDM-TW &  & 2 & 45.01 & 38.15\% &  & 2 & 47.97 & 41.76\% \\
 & AEDM-OR-TW &  & 2 & 52.28 & 28.16\% &  & 2 & 50.40 & 38.81\% \\
& \cellcolor{myblue}UM & \cellcolor{myblue} & \cellcolor{myblue}2 & \cellcolor{myblue}72.77 & \cellcolor{myblue}0.00\% & \cellcolor{myblue} & \cellcolor{myblue}2 & \cellcolor{myblue}82.36 & \cellcolor{myblue}0.00\% \\
\midrule
\multirow{18}{*}{PDRA-TW} & \multirow{2}{*}{Gurobi} & \multirow{9}{*}{2} & 60 & \ding{55} & 100\% & \multirow{9}{*}{3} & 60 & \ding{55} & 100\% \\
 &  &  & 30×60 & 26.01 & 0.27\% &  & 30×60 & 28.52 & 20.40\% \\
 & GC-TW &  & 5 & 12.21 & 53.18\% &  & 11 & 18.21 & 49.17\% \\
 & GC+LS-TW &  & 78 & 19.64 & 24.69\% &  & 241 & 28.14 & 21.46\% \\
 & AEDM-Basic &  & 1 & 23.23 & 10.93\% &  & 1 & 32.42 & 9.52\% \\
 & AEDM-OR &  & 1 & 22.27 & 14.61\% &  & 1 & 28.87 & 19.43\% \\
 & AEDM-TW &  & 1 & 26.03 & 0.19\% &  & 1 & 35.86 & -0.08\% \\
 & AEDM-OR-TW &  & 1 & 25.99 & 0.35\% &  & 1 & 34.51 & 3.68\% \\
& \cellcolor{myblue}UM & \cellcolor{myblue} & \cellcolor{myblue}1 & \cellcolor{myblue}26.08 & \cellcolor{myblue}0.00\% & \cellcolor{myblue} & \cellcolor{myblue}1 & \cellcolor{myblue}35.83 & \cellcolor{myblue}0.00\% \\
& \multirow{2}{*}{Gurobi} & \multirow{9}{*}{4} & 60 & \ding{55} & 100\% & \multirow{9}{*}{5} & 60 & \ding{55} & 100\% \\
 &  &  & 30×60 & 18.43 & 57.80\% &  & 30×60 & 24.21 & 51.93\% \\
 & GC-TW &  & 23 & 15.11 & 65.40\% &  & 31 & 21.21 & 57.89\% \\
 & GC+LS-TW &  & 469 & 30.16 & 30.94\% &  & 958 & 36.24 & 28.06\% \\
 & AEDM-Basic &  &2 & 39.90 & 8.63\% &  & 2 & 46.07 & 8.54\% \\
 & AEDM-OR &  & 2 & 34.14 & 21.82\% &  & 2 & 40.47 & 19.65\% \\
 & AEDM-TW &  & 2 & 42.16 & 3.46\% &  & 2& 49.80 & 1.13\% \\
 & AEDM-OR-TW &  & 2 & 41.38 & 5.24\% &  & 2 & 48.37 & 3.97\% \\
& \cellcolor{myblue}UM & \cellcolor{myblue} & \cellcolor{myblue}2 & \cellcolor{myblue}43.67 & \cellcolor{myblue}0.00\% & \cellcolor{myblue} & \cellcolor{myblue}2 & \cellcolor{myblue}50.37 & \cellcolor{myblue}0.00\% \\
\midrule
\multirow{18}{*}{PDRA-OR-TW} & \multirow{2}{*}{Gurobi} & \multirow{9}{*}{2} & 60 & \ding{55} & 100\% & \multirow{9}{*}{3} & 60 & \ding{55} & 100\% \\
 &  &  & 30×60 & 26.02 & 0.95\% &  & 30×60 & 32.11 & 9.32\% \\
 & GC-OR-TW &  & 7 & 10.18 & 61.25\% &  & 15 & 19.91 & 43.78\% \\
 & GC+LS-OR-TW &  & 71 & 25.14 & 4.30\% &  & 305 & 33.12 & 6.47\% \\
 & AEDM-Basic &  & 1 & 24.11 & 8.22\% &  & 1 & 33.02 & 6.75\% \\
 & AEDM-OR &  & 1 & 23.79 & 9.44\% &  & 1 & 31.69 & 10.51\% \\
 & AEDM-TW &  & 1 & 25.58 & 2.63\% &  & 1 & 33.70 & 4.83\% \\
 & AEDM-OR-TW &  & 1 & 26.10 & 0.65\% &  & 1 & 34.92 & 1.38\% \\
& \cellcolor{myblue}UM & \cellcolor{myblue} & \cellcolor{myblue}1 & \cellcolor{myblue}26.27 & \cellcolor{myblue}0.00\% & \cellcolor{myblue} & \cellcolor{myblue}1 & \cellcolor{myblue}35.41 & \cellcolor{myblue}0.00\% \\
 & \multirow{2}{*}{Gurobi} & \multirow{9}{*}{4} & 60 & \ding{55} & 100\% & \multirow{9}{*}{5} & 60 & \ding{55} & 100\% \\
 &  &  & 30×60 & 25.29 & 42.67\% &  & 30×60 & 25.14 & 50.53\% \\
 & GC-OR-TW &  & 24 & 15.20 & 65.55\% &  & 35 & 14.50 & 71.47\% \\
 & GC+LS-OR-TW &  & 620 & 23.12 & 47.60\% &  & 1104 & 22.18 & 56.36\% \\
 & AEDM-Basic &  & 2 & 40.37 & 8.48\% &  & 2 & 46.98 & 7.56\% \\
 & AEDM-OR &  & 2 & 39.55 & 10.34\% &  & 2 & 46.46 & 8.58\% \\
 & AEDM-TW &  & 2 & 42.00 & 4.78\% &  & 2 & 48.84 & 3.90\% \\
 & AEDM-OR-TW &  & 2 & 43.33 & 1.77\% &  & 2 & 49.00 & 3.58\% \\
& \cellcolor{myblue}UM & \cellcolor{myblue} & \cellcolor{myblue}2 & \cellcolor{myblue}44.11 & \cellcolor{myblue}0.00\% & \cellcolor{myblue} & \cellcolor{myblue}2 & \cellcolor{myblue}50.82 & \cellcolor{myblue}0.00\% \\
\bottomrule
\end{tabular}
\end{table}

\begin{figure}
\centering
\includegraphics[width=0.95\textwidth]{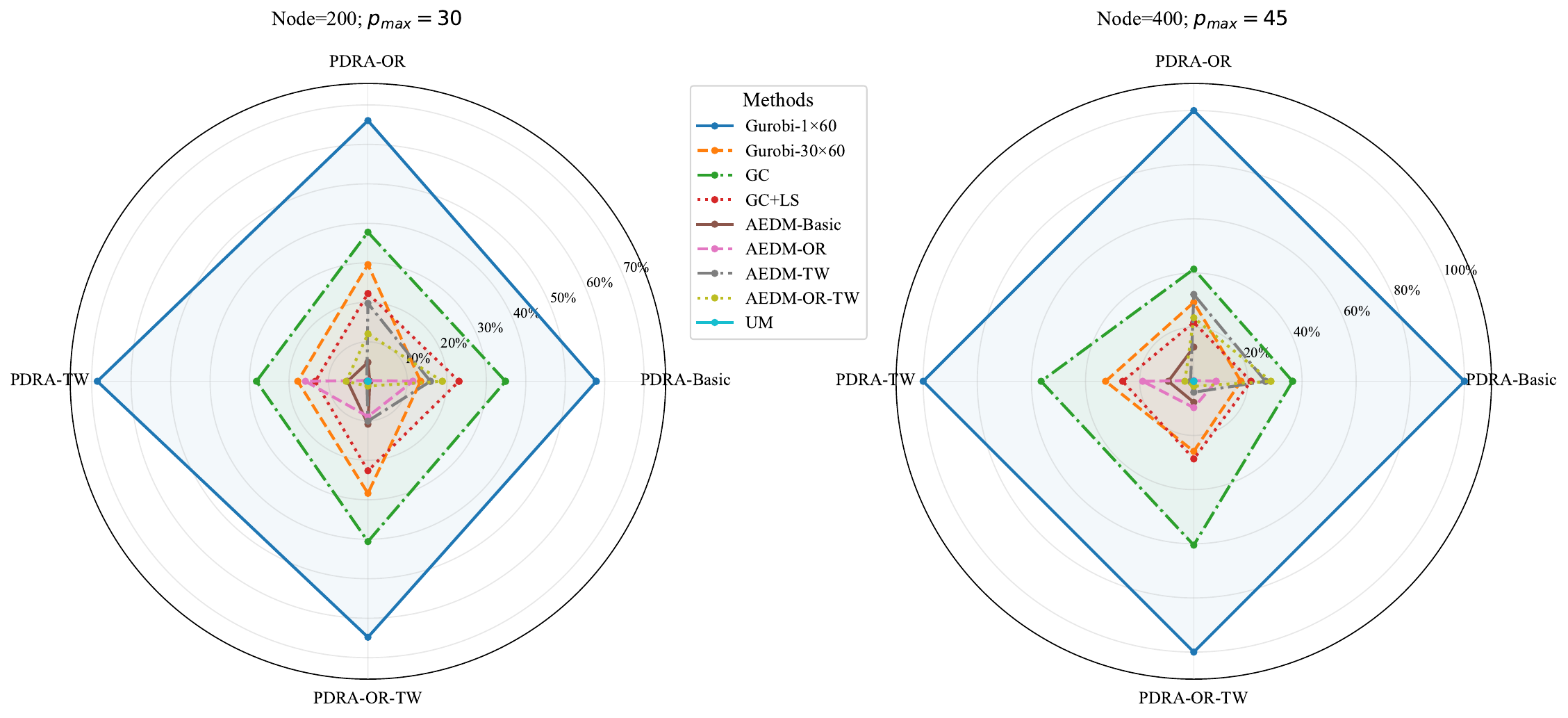}
\caption{Comparative Performance of Methods Across Problem Variants}
\label{fig 5.1}
\end{figure}

\begin{figure}
\centering
\includegraphics[width=0.9\textwidth]{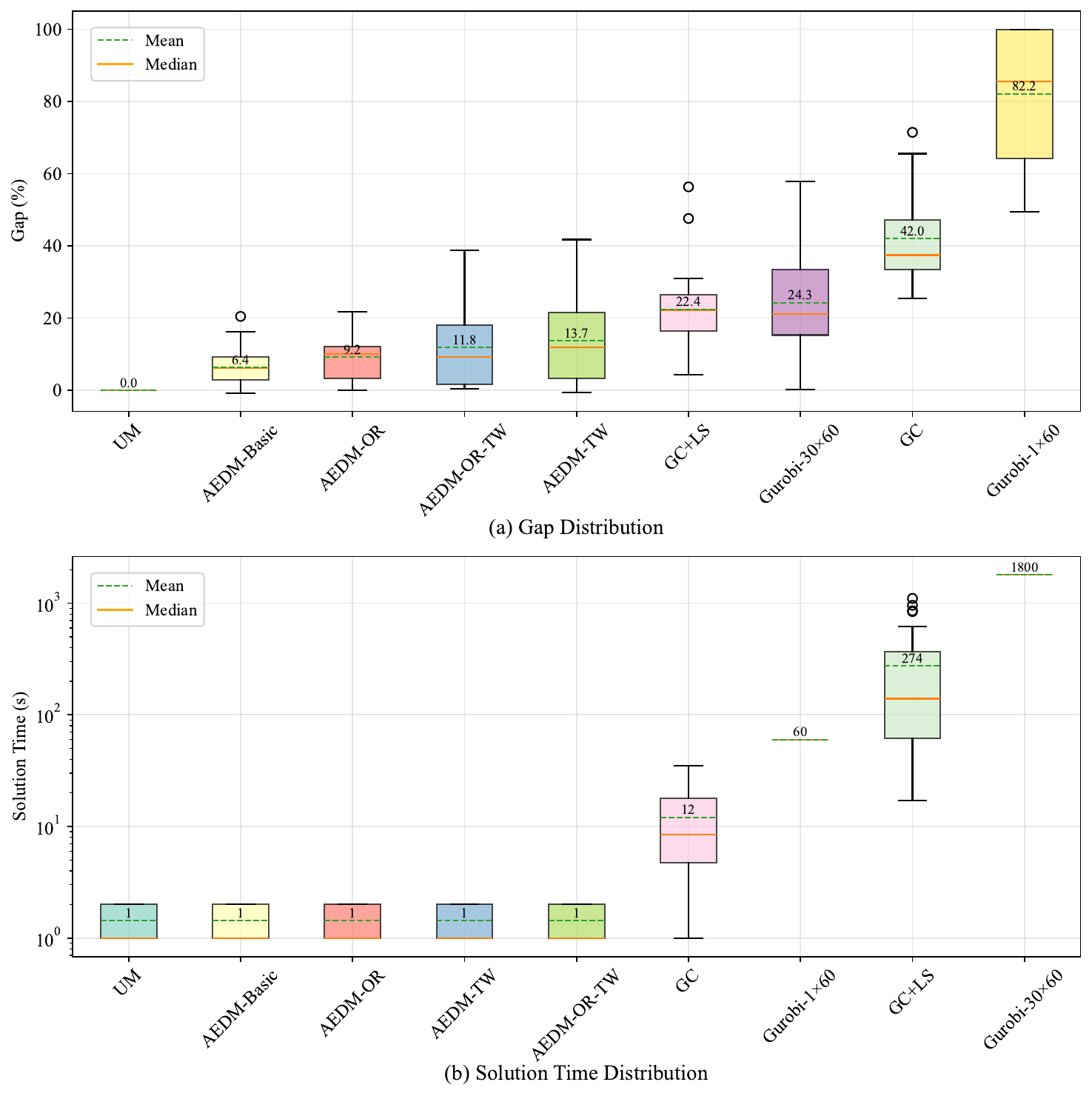}
\caption{Performance Gap and Solution Time Distribution Across Methods}
\label{fig 5.2}
\end{figure}

Tables~\ref{Table 4} and~\ref{Table 5} present the numerical experiments of the comparisons for 200-node and 400-node networks, where the relative performance gap is calculated as $\text{Gap} = (y - y_{\text{other}})/y$, where $y$ denotes the objective value of UM and $y_{\text{other}}$ that of the comparison method. From these results, two key insights emerge:

\begin{enumerate}[label=\alph*)]
\item \textbf{Solution Quality Superiority.} UM consistently achieves the highest objective values across all PDRA variants. For the commercial solver (Gurobi), solution quality gradually improves as the time limit extends; however, even with the longest time limit (1,800 seconds), its final solution quality remains suboptimal compared to UM. In 400-node networks, Gurobi frequently fails to generate feasible solutions within practical time limits (marked by \ding{55} in Table~\ref{Table 5}). For heuristic methods, GC+LS achieves better performance than GC alone through local search improvement, but both remain substantially inferior to UM. For single-task AEDMs, high solution quality is only achieved when the model is specifically trained for the target PDRA variant; when applied to non-matching variants (e.g., using AEDM-Basic for PDRA-OR-TW, which is feasible due to identical model architectures), performance degrades significantly.

\item \textbf{Computational Efficiency.} UM delivers rapid performance, generating valid solutions within 1--2 seconds across all tested network scales and drone fleet sizes. In contrast, Gurobi requires 1,800 seconds, (30 minutes) to reach reasonable-quality solutions, far exceeding the time-critical demands of post-disaster response. Heuristic methods exhibit varying computational times: GC provides rapid construction but suboptimal quality, while GC+LS trades additional computational time (ranging from seconds to minutes depending on problem complexity) for improved solutions that still fall short of UM's performance. Single-task AEDMs also suffer from inefficiency as they require separate models (one for each PDRA variant), leading to increased total training time and necessitating variant-specific deployment, which further complicates real-world application.
\end{enumerate}

Figures~\ref{fig 5.1} and~\ref{fig 5.2} visually corroborate these findings. The radar charts in Figure~\ref{fig 5.1} show UM consistently occupying the outermost (best) performance positions across all parameter settings and problem variants, indicating superior and stable performance under diverse operational constraints. Figure~\ref{fig 5.2} (a) shows that UM consistently outperforms baseline methods in terms of solution quality (total collected information value). Specifically, UM surpasses the best single-task DRL models by 6\%--14\%, heuristic methods (GC+LS) by 22\%--42\%, and Gurobi by 24\%--82\% across tested configurations. Figure~\ref{fig 5.2} (b) demonstrates the computational efficiency advantages: UM maintains rapid performance with solution times of 1--2 seconds, comparable to single-task AEDM, while substantially outperforming heuristic methods (12--274 seconds) and Gurobi (60--1,800 seconds).

\subsection{Sensitivity analysis}

To evaluate the robustness and scalability of the proposed UM, this section conducts sensitivity analyses across three critical dimensions: (i) maximum allowable assessment time, (ii) drone fleet size, and (iii) network scale. These analyses assess performance stability when key parameters deviate from training conditions, demonstrating the model's applicability to diverse post-disaster scenarios. All sensitivity tests are conducted on the PDRA-OR-TW variant, which simultaneously incorporates the open route and time window constraints, thereby representing a complex and practically relevant scenario. The results yield three key insights:
\begin{enumerate}[label=\alph*)]
\item Table~\ref{Table 6} reports performance under varying assessment time limits (20, 30, 40, and 50 minutes) on a fixed 400-node network with four drones. This experiment reflects the time-critical nature of PDRA missions, where available assessment windows may vary by disaster severity or resource constraints. UM maintains superior solution quality across all tested settings, with performance gaps over the best single-task AEDM ranging from approximately $-0.20\%$ to $16.07\%$. Gurobi fails to produce feasible solutions within a 60-second limit across all time windows, requiring full 30-minute runs to return suboptimal results. The GC and GC+LS heuristics remain consistently inferior to UM, although GC+LS achieves noticeable improvements over GC. These findings confirm UM’s robustness to temporal variability in PDRA operations.

\item Table~\ref{Table 7} examines scenarios with 4, 6, 8, and 10 drones on a fixed 400-node network and $p_{\max}=45$ minutes, simulating practical fluctuations in drone availability during disaster response. UM consistently outperforms all baselines, with performance gaps of $1.20\%$--$12.79\%$ relative to the best AEDM model. Gurobi again fails to produce feasible solutions within the 60-second limit for all fleet sizes, highlighting UM’s advantage in time-critical PDRA settings. The GC and GC+LS heuristics show predictable degradation as complexity increases, though GC+LS remains noticeably stronger than GC. UM’s computational time scales linearly with the number of drones (approximately 2--10 seconds), illustrating efficient parallelization capability for multi-drone coordination.

\item Table~\ref{Table 8} evaluates scalability on networks of 400, 600, 800, and 1,000 nodes while keeping $p_{\max}=45$ minutes and the drone fleet size fixed at four drones. UM demonstrates strong scalability, preserving solution-quality superiority across all scales, with performance gaps of $0.17\%$--$11.65\%$ over the best AEDM. The GC and GC+LS heuristics show increasing deterioration as network size grows, reflecting their structural limitations under large-scale PDRA settings. Gurobi becomes completely intractable beyond 400 nodes, failing even with an extended 30-minute limit. Despite increased problem size, UM maintains rapid computational performance (2--10 seconds), underscoring its practical viability for large-scale PDRA applications.
\end{enumerate}

\begin{table}[!]
\scriptsize 
\caption{Sensitivity analysis under varying maximum allowable assessment times ($p_{\max}$) (400 nodes, 4 drones)}
\label{Table 6}
\resizebox{\textwidth}{!}{
\begin{tabular}{ccccccccccccc}
\toprule
& \multicolumn{12}{c}{Maximum allowable assessment times}\\
\cmidrule(lr){2-13}
\multirow{2}{*}{Method} & \multicolumn{3}{c}{20} & \multicolumn{3}{c}{30} & \multicolumn{3}{c}{40} & \multicolumn{3}{c}{50} \\
\cmidrule(lr){2-4} \cmidrule(lr){5-7} \cmidrule(lr){8-10} \cmidrule(lr){11-13}
& Value & Gap & Time (s) & Value & Gap & Time (s) & Value & Gap & Time (s) & Value & Gap & Time (s) \\
\midrule
\multirow{2}{*}{Gurobi} & \ding{55} & 100\% & 60 & \ding{55} & 100\% & 60 & \ding{55} & 100\% & 60 & \ding{55} & 100\% & 60 \\
 & 8.25 & 33.04\% & 30×60 & 15.22 & 40.59\% & 30×60 & 22.40 & 42.04\% & 30×60 & 28.79 & 42.74\% & 30×60 \\
GC-OR-TW & 6.04 & 50.97\% & 9 & 9.89 & 61.40\% & 15 & 13.57 & 64.89\% & 28 & 14.85 & 70.46\% & 34 \\
GC+LS-OR-TW & 8.12 & 34.09\% & 243 & 16.24 & 36.61\% & 426 & 21.19 & 45.18\% & 710 & 32.14 & 36.08\% & 1014 \\
AEDM-Basic & 11.91 & 3.33\% & 1 & 23.73 & 7.38\% & 2 & 34.47 & 10.82\% & 2 & 42.20 & 16.07\% & 2 \\
AEDM-OR & 11.93 & 3.17\% & 1 & 21.82 & 14.83\% & 2 & 34.62 & 10.43\% & 2 & 43.33 & 13.82\% & 2 \\
AEDM-TW & 11.89 & 3.49\% & 1 & 22.59 & 11.83\% & 2 & 34.44 & 10.89\% & 2 & 44.37 & 11.75\% & 2 \\
AEDM-OR-TW & 12.34 & -0.16\% & 1 & 25.67 & -0.20\% & 2 & 37.84 & 2.10\% & 2 & 49.14 & 2.27\% & 2 \\
\rowcolor{myblue}
UM & 12.32 & 0.00\% & 1 & 25.62 & 0.00\% & 2 & 38.65 & 0.00\% & 2 & 50.28 & 0.00\% & 2 \\
\bottomrule
\end{tabular}
}
\end{table}

\begin{table}[!]
\scriptsize 
\caption{Sensitivity analysis under varying drone fleet sizes (400 nodes, $p_{\max}=45$)}
\label{Table 7}
\resizebox{\textwidth}{!}{
\begin{tabular}{ccccccccccccc}
\toprule 
& \multicolumn{12}{c}{Drone fleet sizes}\\
\cmidrule(lr){2-13}
\multirow{2}{*}{Method} & \multicolumn{3}{c}{4} & \multicolumn{3}{c}{6} & \multicolumn{3}{c}{8} & \multicolumn{3}{c}{10} \\
\cmidrule(lr){2-4} \cmidrule(lr){5-7} \cmidrule(lr){8-10} \cmidrule(lr){11-13}
& Value & Gap & Time (s) & Value & Gap & Time (s) & Value & Gap & Time (s) & Value & Gap & Time (s) \\
\midrule
\multirow{2}{*}{Gurobi} & \ding{55} & 100\% & 60 & \ding{55} & 100\% & 60 & \ding{55} & 100\% & 60 & \ding{55} & 100\% & 60 \\
 & 25.29 & 42.67\% & 30×60 & 32.76 & 41.35\% & 30×60 & 38.50 & 40.28\% & 30×60 & 42.80 & 39.45\% & 30×60 \\
GC-OR-TW & 15.20 & 65.54\% & 24 & 21.35 & 61.78\% & 57 & 25.80 & 59.90\% & 95 & 29.50 & 58.35\% & 138 \\
GC+LS-OR-TW & 23.12 & 47.60\% & 620 & 38.45 & 31.17\% & 1501 & 48.20 & 25.24\% & 2485 & 55.80 & 21.19\% & 3580 \\
AEDM-Basic & 40.37 & 8.48\% & 2 & 50.28 & 9.99\% & 3 & 56.80 & 11.83\% & 6 & 62.15 & 12.16\% & 10 \\
AEDM-OR & 39.55 & 10.34\% & 2 & 50.20 & 10.13\% & 3 & 56.50 & 12.31\% & 6 & 61.70 & 12.79\% & 10 \\
AEDM-TW & 42.00 & 4.78\% & 2 & 52.37 & 6.25\% & 3 & 58.95 & 8.47\% & 6 & 64.30 & 9.14\% & 10 \\
AEDM-OR-TW & 43.33 & 1.77\% & 2 & 55.19 & 1.20\% & 3 & 61.85 & 3.98\% & 6 & 67.45 & 4.66\% & 10 \\
\rowcolor{myblue}
UM & 44.11 & 0.00\% & 2 & 55.86 & 0.00\% & 3 & 64.42 & 0.00\% & 6 & 70.78 & 0.00\% & 10 \\
\bottomrule
\end{tabular}
}
\end{table}

\begin{table}[!ht]
\scriptsize 
\caption{Sensitivity analysis under varying network scales ($p_{\max}=45$, 4 drones)}
\label{Table 8}
\resizebox{\textwidth}{!}{ 
\begin{tabular}{ccccccccccccc}
\toprule
& \multicolumn{12}{c}{Network scales}\\
\cmidrule(lr){2-13}
\multirow{2}{*}{Method} & \multicolumn{3}{c}{400} & \multicolumn{3}{c}{600} & \multicolumn{3}{c}{800} & \multicolumn{3}{c}{1000} \\
\cmidrule(lr){2-4} \cmidrule(lr){5-7} \cmidrule(lr){8-10} \cmidrule(lr){11-13}
& Value & Gap & Time (s) & Value & Gap & Time (s) & Value & Gap & Time (s) & Value & Gap & Time (s) \\
\midrule
\multirow{2}{*}{Gurobi} & \ding{55} & 100\% & 60 & \ding{55} & 100\% & 60 & \ding{55} & 100\% & 60 & \ding{55} & 100\% & 60 \\
 & 25.29 & 42.67\% & 30×60 & \ding{55} & 100\% & 30×60 & \ding{55} & 100\% & 30×60 & \ding{55} & 100\% & 30×60 \\
GC-OR-TW & 15.20 & 65.54\% & 24 & 15.94 & 66.77\% & 35 & 17.51 & 66.39\% & 57 & 21.41 & 62.06\% & 87 \\
GC+LS-OR-TW & 23.12 & 47.60\% & 620 & 24.21 & 49.53\% & 941 & 27.54 & 47.13\% & 1431 & 31.24 & 44.64\% & 2104 \\
AEDM-Basic & 40.37 & 8.48\% & 2 & 43.06 & 10.23\% & 4 & 47.24 & 9.31\% & 8 & 49.91 & 11.57\% & 10 \\
AEDM-OR & 39.55 & 10.34\% & 2 & 42.38 & 11.65\% & 4 & 46.22 & 11.27\% & 8 & 50.53 & 10.48\% & 10 \\
AEDM-TW & 42.00 & 4.78\% & 2 & 45.34 & 5.48\% & 4 & 49.00 & 5.93\% & 8 & 53.17 & 5.79\% & 10 \\
AEDM-OR-TW & 43.33 & 1.77\% & 2 & 47.89 & 0.17\% & 4 & 51.66 & 0.83\% & 8 & 55.83 & 1.08\% & 10 \\
\rowcolor{myblue}
UM & 44.11 & 0.00\% & 2 & 47.97 & 0.00\% & 4 & 52.09 & 0.00\% & 8 & 56.44 & 0.00\% & 10 \\
\bottomrule
\end{tabular}
}
\end{table}

The sensitivity analyses collectively validate UM's robustness across critical operational parameters, confirming its suitability for deployment in diverse post-disaster environments where parameter variations are inevitable. Across all tested dimensions of temporal constraints, fleet sizes, and network scales, UM demonstrates consistent performance superiority and computational efficiency, confirming the model's practical reliability for time-sensitive emergency response operations where both solution quality and computational speed are paramount. Furthermore, the comparisons with both heuristic methods and exact optimization highlight the fundamental limitations of traditional approaches: heuristics require extensive domain expertise and variant-specific design yet achieve suboptimal solutions, while exact methods suffer from computational intractability on realistic problem scales. UM successfully overcomes both limitations through learned multi-task representations that generalize effectively across diverse operational scenarios.

\subsection{Finetuning}  
\label{Finetuning}  

To evaluate the proposed lightweight finetuning mechanism for incorporating previously unseen attributes, experiments are conducted on 400-node networks with 2 depots, each deploying 3 drones under a maximum assessment time constraint of $p_{max} = 45$ minutes. The MD attribute, deliberately excluded from initial UM training, serves as the representative unseen attribute to assess adaptation capabilities. The finetuning mechanism leverages efficient adapter layers through parameter augmentation with zero-initialized entries, requiring only minimal modifications to the input embedding and decoder's context embedding while preserving the valuable multi-task knowledge encoded in the original parameters. Five approaches are compared: (i) Gurobi, representing exact optimization; (ii) GC and GC+LS heuristics, representing heuristic optimization with their variant-specific implementations for MD-integrated problems (GC-Variant and GC+LS-Variant); (iii) single-task AEDM variants (AEDM-MD, AEDM-OR-MD, AEDM-TW-MD, and AEDM-OR-TW-MD), each trained from scratch for 200 epochs; (iv) UM-Zero-Shot, where the pre-trained UM is directly applied to MD problems without adaptation; and (v) UM-10-Epochs, which applies the lightweight adapter mechanism with only 10 epochs of finetuning. The results are summarized in Table~\ref{Table 9}.

Compared with Gurobi, UM-10-Epochs shows advantages in both solution quality and efficiency. Gurobi often fails to produce feasible solutions within the 60-second limit and requires full 30 minutes to obtain suboptimal outcomes, whereas UM-10-Epochs consistently generates high-quality solutions in just a few seconds. For heuristic methods, GC+LS achieves better performance than GC alone through local search improvement, but both remain substantially inferior to UM-10-Epochs. When compared with single-task AEDM variants, UM-10-Epochs achieves higher solution quality despite AEDMs undergoing full training on each MD-integrated problem. This highlights that multi-task knowledge transfer, coupled with lightweight finetuning, allows UM to learn new attribute-specific knowledge at only a fraction of the training cost, avoiding the inefficiency of maintaining multiple specialized models. Relative to UM-Zero-Shot, UM-10-Epochs eliminates the moderate performance gap that arises when the MD attribute is directly introduced without adaptation. The short 10-epoch finetuning is sufficient to fully close this gap, yielding better performance across all MD-integrated PDRA variants. Overall, these results confirm that even minimal finetuning enables UM to reconcile pre-trained knowledge with new operational constraints, validating the effectiveness of the adapter mechanism. UM thus proves particularly valuable for rapid adaptation in evolving disaster response scenarios, where new operational requirements may emerge with limited preparation time.

\begin{table}[]
\footnotesize
\caption{Performance Comparison of Methods for MD-Integrated PDRA Variants}
\label{Table 9}
\resizebox{\textwidth}{!}{\begin{tabular}{ccccccccccccc}
\toprule
\multirow{2}{*}{Method} & \multicolumn{3}{c}{AEDM-MD} & \multicolumn{3}{c}{AEDM-OR-MD} & \multicolumn{3}{c}{AEDM-TW-MD} & \multicolumn{3}{c}{AEDM-OR-TW-MD} \\
\cmidrule(lr){2-4} \cmidrule(lr){5-7} \cmidrule(lr){8-10}  \cmidrule(lr){11-13}
 & Value & Gap & Time (s)& Value & Gap & Time (s)& Value & Gap & Time (s)& Value & Gap & Time (s)\\
 \midrule
\multirow{2}{*}{Gurobi} & \ding{55} & 100\% & 60 & \ding{55} & 100\% & 60 & \ding{55} & 100\% & 60 & \ding{55} & 100\% & 60 \\
 & 53.24 & 25.24\% & 30×60 & 61.84 & 26.60\% & 30×60 & 41.45 & 19.06\% & 30×60 & 52.16 & 16.74\% & 30×60 \\
GC-Variant & 41.12 & 42.24\% & 41 & 42.39 & 49.68\% & 40 & 32.56 & 36.42\% & 58 & 38.56 & 38.46\% & 61 \\
GC+LS-Variant & 51.23 & 28.06\% & 924 & 58.45 & 30.62\% & 901 & 38.45 & 24.92\% & 1334 & 50.14 & 19.97\% & 1421 \\
AEDM-MD & 70.25 & 1.35\% & 3 & 69.55 & 17.45\% & 3 & 39.47 & 22.93\% & 4 & 50.85 & 18.83\% & 4 \\
AEDM-OR-MD & 59.19 & 16.88\% & 3 & 83.03 & 1.45\% & 3 & 42.52 & 16.97\% & 4 & 51.87 & 17.21\% & 4 \\
AEDM-TW-MD & 57.38 & 19.42\% & 3 & 62.58 & 25.72\% & 3 & 49.68 & 2.99\% & 4 & 53.25 & 15.00\% & 4 \\
AEDM-OR-TW-MD & 55.83 & 21.60\% & 3 & 66.25 & 21.36\% & 3 & 41.41 & 19.14\% & 4 & 60.53 & 3.38\% & 4 \\
UM-Zero-Shot & 57.92 & 18.66\% & 3 & 66.48 & 21.09\% & 3 & 39.54 & 22.79\% & 4 & 52.25 & 16.60\% & 4 \\
\rowcolor{myblue}
UM-10-Epochs & 71.21 & 0.00\% & 3 & 84.25 & 0.00\% & 3 & 51.21 & 0.00\% & 4 & 62.65 & 0.00\% & 4 \\
\bottomrule 
\end{tabular}}
\end{table}

\subsection{Real-world application}
\label{Real-world application}

\begin{figure}
\centering
\includegraphics[width=0.6\textwidth]{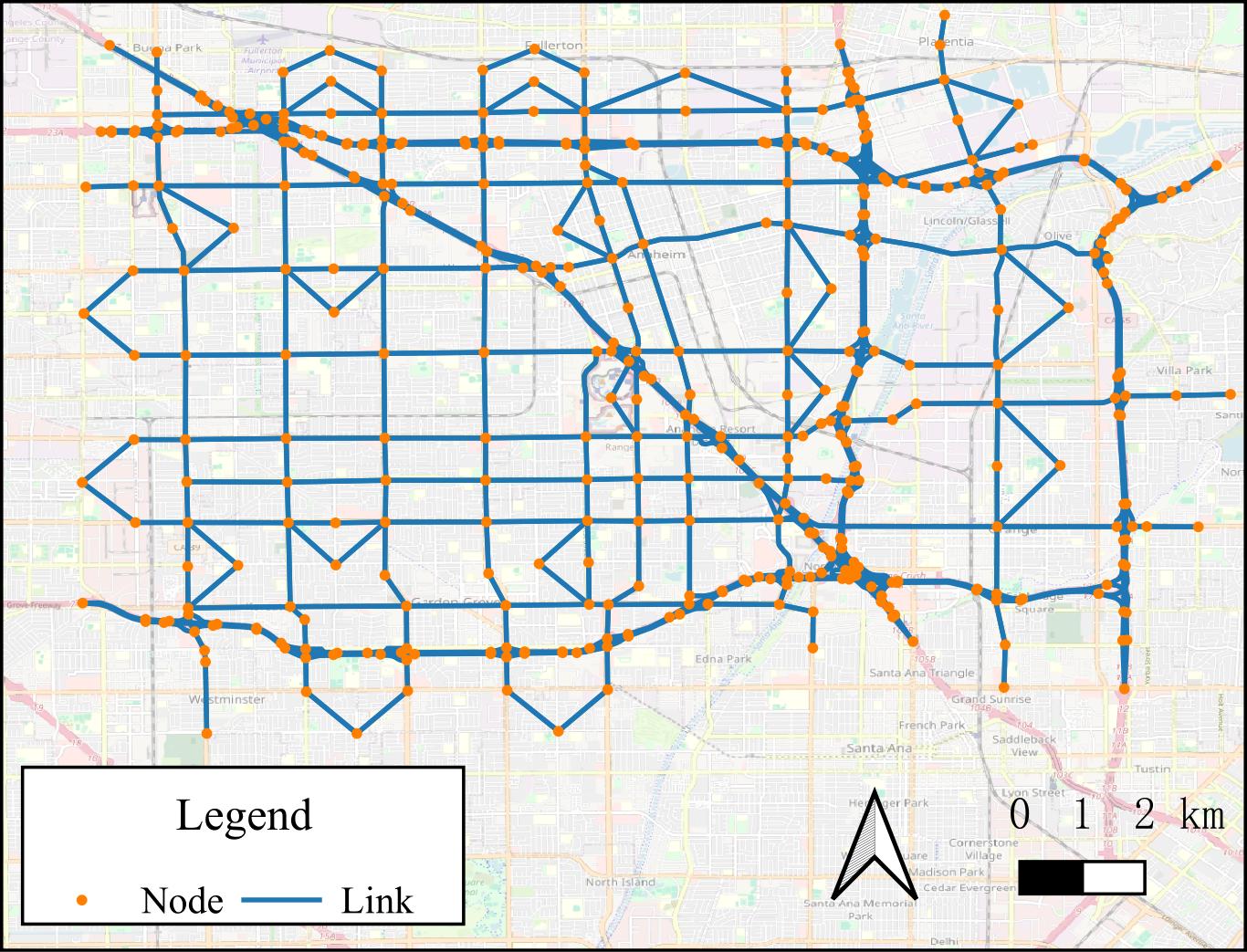}
\caption{Real-World Anaheim Transportation Network}
\label{fig 5.3}
\end{figure}

To further validate the practical applicability of the proposed UM in real-world post-disaster scenarios, experiments are conducted on the publicly available Anaheim transportation network (as shown in Figure~\ref{fig 5.3}), with data sourced from \url{https://github.com/bstabler/TransportationNetworks}. This network simulates the complexity of a road system typical of large-scale PDRAs. The original network (416 nodes and 914 links) is transformed into a 1330-node structure (including 914 artificial nodes for road links) using the transformation method detailed in Section \ref{Network transformation}. The experimental setup includes 10 drones, and the maximum assessment time $p_{\max}$ is set to 45 minutes. Four benchmark approaches are evaluated: (i) UM, solving all 8 PDRA variants in a unified manner; (ii) single-task AEDMs, with 8 separate models trained for each variant; (iii) GC and GC+LS heuristics, with 8 variant-specific implementations; and (iv) Gurobi, with the time limit extended to 10,000 seconds to account for increased problem complexity. 

\begin{figure}
\centering
\includegraphics[width=\textwidth]{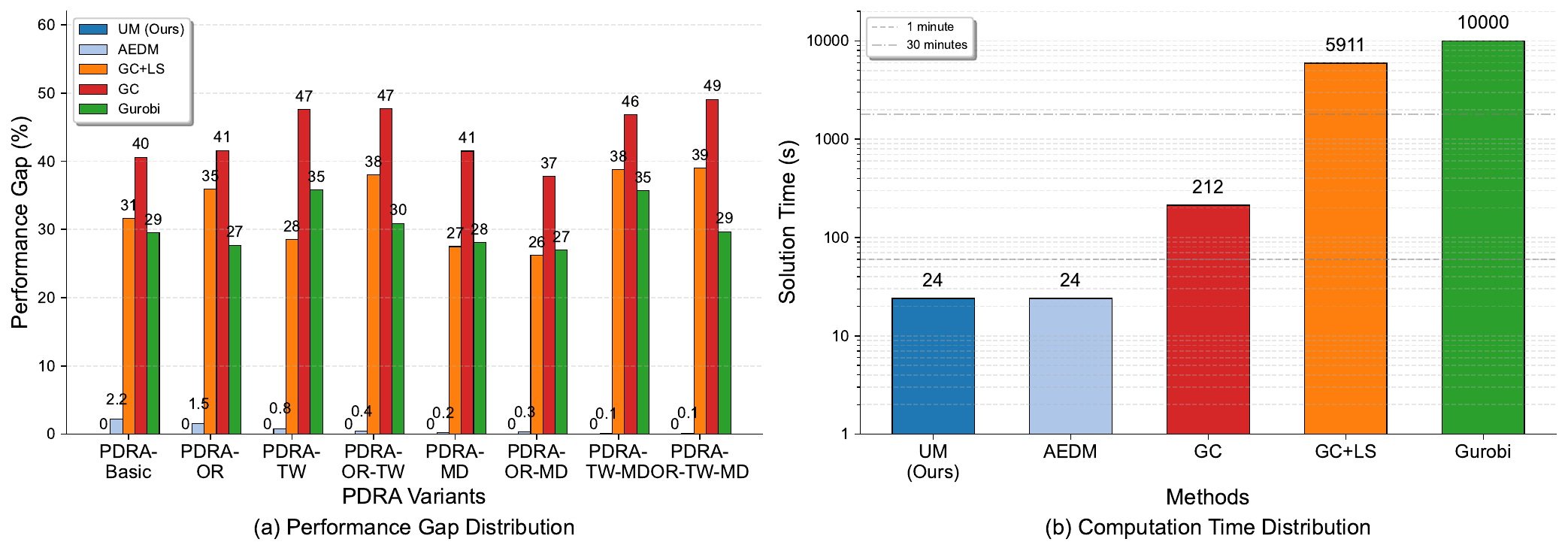}
\caption{Solution Quality and Computational Efficiency on Anaheim Network}
\label{fig 5.4}
\end{figure}

Figure~\ref{fig 5.4} presents the comparative results. Two key observations emerge from the real-world network evaluation. As shown in Figure~\ref{fig 5.4} (a), UM consistently achieves superior solution quality compared to all baseline methods on the large-scale Anaheim network. UM outperforms the best single-task AEDM, demonstrating that the unified multi-task learning approach maintains its advantage even on realistic network topologies. Compared to heuristic methods, UM shows substantial improvements over both GC+LS and GC. These results confirm that hand-crafted heuristics, despite incorporating local search improvements, remain fundamentally limited by their reliance on predefined rules that fail to capture the complex patterns inherent in real-world road networks. Most notably, Gurobi exhibits the poorest performance among all methods. Even with extended time limits, Gurobi frequently fails to generate high-quality feasible solutions for the large-scale network, highlighting the computational intractability of exact optimization methods for real-world large-scale PDRA applications. Figure~\ref{fig 5.4} (b) summarizes the average computational time of these methods for solving the 8 variants and illustrates the dramatic computational efficiency advantages of UM for real-world deployment. UM maintains rapid performance across all PDRA variants. In stark contrast, Gurobi requires extended time limits to return suboptimal solutions, making it impractical for time-critical disaster response scenarios where decisions must be made within minutes. Heuristic methods demonstrate intermediate performance: GC provides rapid construction but at the cost of poor solution quality, while GC+LS achieves moderate improvements through local search but requires substantially longer computation time. Notably, despite requiring separate models for each variant and higher total parameter counts, single-task AEDMs achieve comparable inference speed to UM due to the inherently rapid nature of neural network forward passes. However, UM's unified architecture offers critical advantages in deployment flexibility and training efficiency. By requiring only one model with significantly fewer parameters and reduced training time, UM maintains equivalent rapid performance and superior solution quality compared to multiple specialized models. These findings confirm UM's scalability and robustness under realistic network complexity. The model successfully handles the irregular topology, varied connectivity patterns, and large-scale structure of real-world transportation networks while maintaining both solution quality superiority and rapid computational performance. The results validate UM's practical viability for rapid deployment in time-critical PDRA applications, where both high-quality routing decisions and rapid response times are essential for effective disaster response operations.

\section{Conclusion}
\label{Conclusion}

This study addresses the critical challenge of efficient drone routing for PDRA, where existing methods suffer from computational inefficiencies and lack adaptability to diverse operational scenarios. Traditional approaches, including exact optimization methods, heuristic 
optimization methods, and single-task deep learning models, require separate solutions for each PDRA variant, resulting in substantial computational overhead and deployment complexity. We propose a UM that consolidates all PDRA variants within a single model through multi-task learning. UM employs a modern transformer architecture incorporating RMS normalization, pre-normalization configuration, FA, and SGLUFFN to enhance computational efficiency and solution quality. The unified approach eliminates the need for variant-specific models while leveraging shared knowledge across different operational constraints. Experimental evaluation demonstrates substantial improvements over existing approaches. UM achieves an 8-fold reduction in training time (24 vs. 192 hours) and model parameters (1.3M vs. 10.4M) compared to training separate models for each variant. Solution quality consistently surpasses single-task DRL methods by 6--14\%, heuristic algorithms by 22--42\% and commercial solvers by 24--82\%, while maintaining rapid performance (1--10 seconds) regardless of network scale or operational complexity. Sensitivity analyses confirm robust performance across varying drone fleet sizes, assessment time constraints, and network scales up to 1,000 nodes. The lightweight adapter mechanism enables efficient incorporation of previously unseen attributes through minimal parameter augmentation and brief finetuning, demonstrating practical adaptability to evolving disaster response requirements. 

Future research can proceed in several directions. First, inspired by the scaling laws observed in LLM, where performance improves as parameter size grows \citep{achiam2023gpt,touvron2023llama,guo2025deepseek}, we plan to investigate whether increasing the parameter scale of UM (from millions to potentially billions) can yield further gains in solution quality, generalization, and adaptability. Second, beyond PDRA, the unified multi-task learning framework could be applied to broader classes of neural combinatorial optimization problems in emergency logistics, transportation planning, and other time-critical domains, contributing to the development of general-purpose decision-making models for complex real-world systems.





\bibliographystyle{cas-model2-names}
\bibliography{cas-refs}

\newpage
\appendix

\section{Model component details}
\label{A. Model component details}

This appendix provides mathematical formulations for the key architectural components in Section \ref{Development of the unified model}.

\subsubsection*{RMS}
\label{RMS}

RMS normalization, originally introduced by \cite{zhang2019root}, provides improved training stability and reduced computational overhead compared to traditional instance normalization. In our encoder architecture, RMS normalization replaces instance normalization as described in Equation~\eqref{equation 3}. For a given input vector $\mathbf{x} \in \mathbb{R}^d$, where $d$ represents the embedding dimension, RMS normalization is computed as:

\begin{equation}
\label{A1}
\text{RMS}(\mathbf{x}) = \frac{\mathbf{x}}{\sqrt{\frac{1}{d}\sum_{i=1}^{d}x_i^2 + \epsilon}} \odot \mathbf{g} \tag{A1}
\end{equation}
where $\odot$ denotes element-wise multiplication, $\mathbf{g} \in \mathbb{R}^d$ is a learnable scaling parameter vector, and $\epsilon \approx 10^{-8}$ is a small constant introduced for numerical stability. Unlike instance normalization, RMS normalization eliminates the mean centering operation, thereby reducing computational complexity while maintaining normalization effectiveness. This modification is particularly beneficial for large-scale routing problems where computational efficiency is paramount.

\subsubsection*{FA}
\label{FA}

FA optimizes memory usage and computational efficiency in attention mechanisms without sacrificing numerical accuracy \citep{dao2022flashattention, dao2023flashattention}, as referenced in Equations~\eqref{equation 4} and~\eqref{equation 7}. The standard MHA computation is given by:

\begin{equation}
\label{A2}
\text{Attention}(\mathbf{Q}, \mathbf{K}, \mathbf{V}) = \text{softmax}\left(\frac{\mathbf{Q}\mathbf{K}^T}{\sqrt{d_k}}\right)\mathbf{V} \tag{A2}
\end{equation}
where $\mathbf{Q} \in \mathbb{R}^{n \times d_q}$, $\mathbf{K} \in \mathbb{R}^{n \times d_k}$, and $\mathbf{V} \in \mathbb{R}^{n \times d_v}$ denote the query, key, and value matrices, respectively. Here, $n$ represents the sequence length, which corresponds to the number of nodes in our drone routing context, and $d_q$, $d_k$, and $d_v$ stand for the dimensions of the query, key, and value vectors, respectively. FA reformulates this computation using memory-efficient tiling and recomputation strategies:

\begin{equation}
\label{A3}
\text{FA}(\mathbf{Q}, \mathbf{K}, \mathbf{V}) = \text{BlockwiseAttention}(\mathbf{Q}_{[1:B_q]}, \mathbf{K}_{[1:B_k]}, \mathbf{V}_{[1:B_v]}) \tag{A3}
\end{equation}
where $B_q$, $B_k$, and $B_v$ represent block sizes optimized for the memory hierarchy of the underlying hardware. This approach significantly reduces memory complexity from $\mathcal{O}(n^2)$ to $\mathcal{O}(n)$ for sequence length $n$, enabling efficient processing of large road networks with thousands of nodes.

\subsubsection*{SGLUFFN}
\label{SGLUFFN}

The SGLUFFN component, as defined in Equation~\eqref{equation 5}, enhances the FFN's expressiveness through an advanced gating mechanism. The Swish Gated Linear Unit activation function combines the benefits of gating with smooth activation properties. For input vector $\mathbf{x} \in \mathbb{R}^d$, the SGLUFFN transformation is computed as:

\begin{equation}
\label{A4}
\text{SGLUFFN}(\mathbf{x}) = \text{Swish}(\mathbf{x}\mathbf{W}_1 + \mathbf{b}_1) \odot (\mathbf{x}\mathbf{W}_2 + \mathbf{b}_2) \tag{A4}
\end{equation}
where the Swish activation function is defined as:

\begin{equation}
\label{A5}
\text{Swish}(\mathbf{z}) = \mathbf{z} \odot \sigma(\mathbf{z}) \tag{A5}
\end{equation}
and $\sigma(\cdot)$ denotes the sigmoid function $\sigma(z) = \frac{1}{1 + e^{-z}}$. The learnable parameters include transformation matrices $\mathbf{W}_1, \mathbf{W}_2 \in \mathbb{R}^{d \times d_{f}}$ and bias vectors $\mathbf{b}_1, \mathbf{b}_2 \in \mathbb{R}^{d_{f}}$, where $d_{f}$ represents the hidden dimension of the FFN. The gating mechanism ($\odot$) enables the network to selectively emphasize or suppress different features, improving representational capacity compared to traditional ReLU-based FFN. This enhancement is particularly valuable for capturing complex non-linear relationships in drone routing optimization.

\subsubsection*{SHA}
\label{SHA}

SHA, as employed in Equation~\eqref{equation 8}, simplifies the MHA mechanism for decoder operations where reduced complexity is beneficial while maintaining essential attention capabilities. For query vector $\mathbf{q} \in \mathbb{R}^{d_q}$, key matrix $\mathbf{K} \in \mathbb{R}^{n \times d_k}$, and value matrix $\mathbf{V} \in \mathbb{R}^{n \times d_v}$:

\begin{equation}
\label{A6}
\text{SHA}(\mathbf{q}, \mathbf{K}, \mathbf{V}) = \text{softmax}\left(\frac{\mathbf{q}\mathbf{K}^T}{\sqrt{d_k}}\right)\mathbf{V}. \tag{A6}
\end{equation}
In our decoder architecture, SHA processes the context embedding $\mathbf{h}_c^{(t)}$ with node representations to compute selection probabilities for the next node in the route.

\subsubsection*{Softmax}
\label{Softmax}

The softmax function, as employed in Equation~\eqref{equation 9}, converts raw attention scores into probability distributions for node selection during the decoding process. For a vector of logits $\mathbf{u} = [u_0, u_1, \ldots, u_n] \in \mathbb{R}^{n+1}$, where $u_i$ represents the unnormalized score for selecting node $i$:

\begin{equation}
\label{A7}
\text{Softmax}(\mathbf{u}) = \frac{\exp(u_i)}{\sum_{j=0}^{n}\exp(u_j)}. \tag{A7}
\end{equation}

\section{Ablation study}  
\label{B. Ablation study}  

We evaluate each architectural enhancement in the modern transformer encoder through controlled ablation experiments. Five variants are tested by replacing individual components with traditional alternatives while maintaining all others: (1) Base Group uses classical transformer with post-normalization, instance normalization, ReLU-based FFN, and standard attention; (2) -RMS Group replaces RMS normalization with instance normalization; (3) -PreNorm Group uses post-normalization; (4) -SGLUFFN Group replaces SGLUFFN with ReLU-based FFN; (5) -FA Group uses standard attention.

\begin{figure}
\centering
\includegraphics[width=0.7\textwidth]{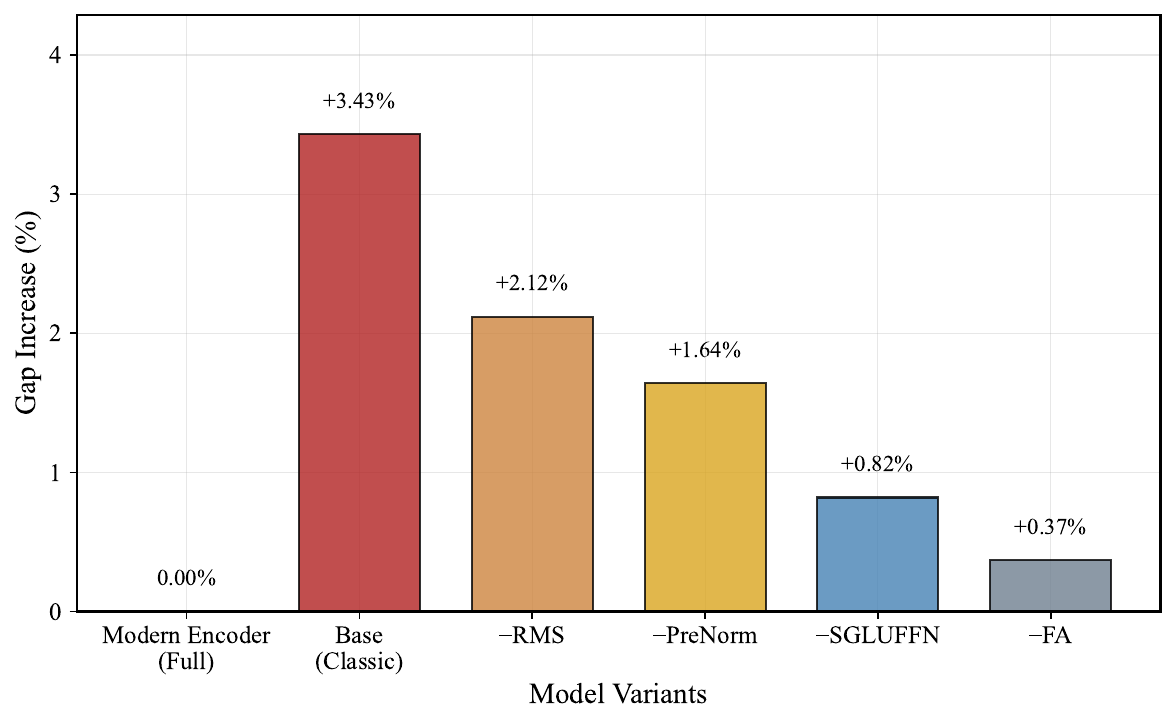}
\caption{Ablation Study Results Showing Performance Gaps for Different Architectural Variants}
\label{fig C1}
\end{figure}

Performance results measured by relative performance gap compared to the full UM reveal significant component contributions. Figure \ref{fig C1} shows the ablation results across different problem instances. The Base Group with traditional components achieves the poorest performance, demonstrating substantial performance gaps. Individual component removals show varying impacts: removing RMS normalization (-RMS) causes moderate performance degradation, while removing pre-normalization (-PreNorm) leads to more significant deterioration. The SGLUFFN activation function proves particularly critical, as its removal (-SGLUFFN) results in notable performance drops. FA removal (-FA) shows the least impact among individual components, though still measurable. These results validate our architectural choices and demonstrate that each modern component contributes meaningfully to the overall model performance.

\section{Mathematical formulation}
\label{C. Mathematical formulation}

This appendix presents the mathematical formulation for the PDRA problem, including attributes for different operational variants. The formulation serves as the basis for commercial solver implementation using Gurobi.

\subsubsection*{Basic model formulation}
\label{Basic model formulation}

Let $x_{ij}^k$ be the binary decision variable indicating whether drone $k$ traverses link $(i,j)$ for assessment, where $x_{ij}^k = 1$ if drone $k$ flies the link, and $x_{ij}^k = 0$ otherwise. Let $f_{ij}^k \geq 0$ be the continuous flow variable representing the flow carried by drone $k$ on link $(i,j)$. The PDRA-Basic model is formulated as \citep{Gong2025Deep}:

\begin{align}
\centering
\label{C1}
& \max \quad \sum_{k \in K} \sum_{p \in \mathcal{P}} c_p \left( \sum_{j:(p,j) \in \bar{A}} x_{pj}^k \right)  \tag{C1} \\
\textbf{s.t.} \quad
\label{C2}
& \sum_{k \in K} \sum_{j:(p,j) \in \bar{A}} x_{pj}^k \leq 1 \quad \forall p \in \mathcal{P} \tag{C2} \\
\label{C3}
& \sum_{j \in \bar{N}} x_{ji}^k  = \sum_{j \in \bar{N}} x_{ij}^k \quad  \forall i \in \bar{N}, \forall k \in K \tag{C3} \\
\label{C4}
& \sum_{j \in \bar{N}} x_{oj}^k  = \sum_{i \in \bar{N}} x_{io}^k  = 1  \quad  \forall k \in K \tag{C4} \\
\label{C5}
& x_{pj}^k + x_{jp}^k \leq 1 \quad \forall j \in \bar{N}, \forall p \in \mathcal{P}, \forall k \in K \tag{C5} \\
\label{C6}
& \sum_{(i,j) \in \bar{A}} t_{ij} x_{ij}^k \leq Q \quad \forall k \in K \tag{C6} \\
\label{C7}
& \max_{k \in K} \left\{ \sum_{(i,j) \in \bar{A}} t_{ij} x_{ij}^k \right\} \leq p_{\max} \tag{C7}\\
\label{C8}
& \sum_{j:(o,j) \in \bar{A}} f_{oj}^k - \sum_{j:(j,o) \in \bar{A}} f_{jo}^k = \sum_{p \in \mathcal{P}} \sum_{j:(p,j) \in \bar{A}} x_{pj}^k \quad \forall k \in K \tag{C8} \\
\label{C9}
& \sum_{j:(j,i) \in \bar{A}} f_{ji}^k - \sum_{j:(i,j) \in \bar{A}} f_{ij}^k = \sum_{j:(i,j) \in \bar{A}} x_{ij}^k \quad \forall i \in \mathcal{P}, \forall k \in K \tag{C9} \\
\label{C10}
& \sum_{j:(j,i) \in \bar{A}} f_{ji}^k - \sum_{j:(i,j) \in \bar{A}} f_{ij}^k = 0 \quad \forall i \in N \setminus \{o\}, \forall k \in K \tag{C10} \\
\label{C11}
& f_{ij}^k \leq |\mathcal{P}| \cdot x_{ij}^k \quad \forall (i,j) \in \bar{A}, \forall k \in K \tag{C11} \\
\label{C12}
& x_{ij}^k \in \{0,1\} \quad  \forall (i,j) \in \bar{A}, \forall k \in K  \tag{C12} \\
\label{C13}
& f_{ij}^k \geq 0 \quad \forall (i,j) \in \bar{A}, \forall k \in K \tag{C13}
\end{align}

where:
\begin{itemize}
\item Equation~\eqref{C1}: Maximize total information value collected from artificial nodes $\mathcal{P}$
\item Equation~\eqref{C2}: Each artificial node visited by at most one drone (no redundant assessment)
\item Equation~\eqref{C3}: Flow conservation constraint ensuring path continuity
\item Equation~\eqref{C4}: Each drone starts and ends at depot $o$
\item Equation~\eqref{C5}: Prevent simultaneous bidirectional traversal of artificial nodes
\item Equation~\eqref{C6}: Flight time constraint within battery limit $Q$
\item Equation~\eqref{C7}: Maximum allowable assessment time constraint within limit $p_{\max}$
\item Equation~\eqref{C8}: Depot flow balance constraint with net outflow equal to number of visited artificial nodes
\item Equation~\eqref{C9}: Flow conservation at artificial nodes with unit consumption for assessment completion
\item Equation~\eqref{C10}: Flow conservation at original network nodes allowing revisits
\item Equation~\eqref{C11}: Flow capacity constraint linking flow variables to routing decisions
\item Equations~\eqref{C12}--\eqref{C13}: Variable domain constraints
\end{itemize}

\subsubsection*{OR attribute}
\label{OR attribute}

For the open route configuration where drones do not return to depot:

\begin{equation}
\label{C14}
t_{io} = 0 \quad \forall i \in \bar{N}. \tag{C14}
\end{equation}

This modification eliminates return-to-depot time costs, allowing drones to terminate routes at any feasible location.

\subsubsection*{TW attribute}
\label{TW attribute}

\textbf{Additional Variables:}
\begin{itemize}
\item $a_i^k$: Arrival time of drone $k$ at node $i$
\item $l_i$: Latest allowable arrival time at node $i$
\end{itemize}

\textbf{Additional Constraints:}
\begin{align}
\label{C15}
& a_i^k \leq l_i \quad \forall i \in \bar{N}, \forall k \in K \tag{C15} \\
\label{C16}
& a_j^k \geq a_i^k + t_{ij} - M(1 - x_{ij}^k) \quad \forall (i,j) \in \bar{A}, \forall k \in K \tag{C16} \\
\label{C17}
& a_o^k = 0 \quad \forall k \in K \tag{C17} \\
\label{C18}
& a_i^k \geq 0 \quad \forall i \in \bar{N}, \forall k \in K \tag{C18}
\end{align}
where Equation~\eqref{C15} enforces time window compliance, Equation~\eqref{C16} ensures time continuity along paths, Equation~\eqref{C17} sets depot departure time, and Equation~\eqref{C18} maintains temporal feasibility.

\subsubsection*{MD attribute}
\label{MD attribute}

\textbf{Additional Variables:}
\begin{itemize}
\item $z_o^k$: Binary variable indicating drone $k$ assignment to depot $o \in \mathcal{D}$
\item $\mathcal{D}$: Set of origin depots
\item $\delta_o$: Maximum number of drones available at depot $o$
\end{itemize}

\textbf{Additional Constraints:}
\begin{align}
\label{C19}
& \sum_{o \in \mathcal{D}} z_o^k \leq 1 \quad \forall k \in K \tag{C19} \\
\label{C20}
& \sum_{j \in \bar{N}} x_{oj}^k \leq z_o^k \quad \forall o \in \mathcal{D}, \forall k \in K \tag{C20} \\
\label{C21}
& \sum_{k \in K} z_o^k \leq \delta_o \quad \forall o \in \mathcal{D} \tag{C21} \\
\label{C22}
& \sum_{j \in \bar{N}} x_{jo}^k = z_o^k \quad \forall d = o \in \mathcal{D}, \forall k \in K \tag{C22} \\
\label{C23}
& z_o^k \in \{0,1\} \quad \forall o \in \mathcal{D}, \forall k \in K \tag{C23}
\end{align}
where Equation~\eqref{C19} assigns each drone to at most one depot, Equation~\eqref{C20} links depot assignment to departure decisions, Equation~\eqref{C21} limits the number of drones at each depot, Equation~\eqref{C22} ensures return to assigned depot, and Equation~\eqref{C23} defines binary depot assignment variables.

\subsubsection*{Implementation notes}
\label{Implementation notes}

\textbf{Model Variants:} The eight PDRA variants are obtained by combining the attributes:
\begin{itemize}
\item PDRA-Basic: Equations~\eqref{C1}--\eqref{C13}
\item PDRA-OR: Add Equation~\eqref{C14}
\item PDRA-TW: Add Equations~\eqref{C15}--\eqref{C18}
\item PDRA-OR-TW: Add Equations~\eqref{C14}--\eqref{C18}
\item PDRA-MD: Add Equations~\eqref{C19}--\eqref{C23}
\item PDRA-OR-MD: Add Equations~\eqref{C14}, \eqref{C19}--\eqref{C23}
\item PDRA-TW-MD: Add Equations~\eqref{C15}--\eqref{C23}
\item PDRA-OR-TW-MD: Add all attribute equations.
\end{itemize}

\section{Heuristic algorithms}
\label{Heuristic algorithms}

To enable extensive performance comparison with the proposed UM, we develop benchmark heuristic algorithms for PDRA variants. While dedicated heuristics could theoretically be designed for each of the eight variants, such an approach would require substantial domain expertise and algorithmic redesign for each configuration, precisely the limitations that motivate our unified learning framework. Therefore, we adopt a representative approach: we implement a two-phase heuristic for the PDRA-TW variant as an exemplar, demonstrating the complexity inherent in traditional optimization methods even for a single variant.

The PDRA-TW variant is selected as the representative case because it encompasses critical operational constraints (time windows) that substantially complicate routing decisions compared to PDRA-Basic. Building upon the classic constructive-improvement paradigm widely employed in combinatorial optimization \citep{kobeaga2018efficient, yang2025heuragenix}, our heuristic must accommodate PDRA-specific structural requirements: (1) artificial nodes representing road links are mutually exclusive and cannot be revisited; (2) these artificial nodes connect only to designated original nodes rather than forming a fully connected subgraph; (3) temporal feasibility constraints must be strictly enforced; and (4) original nodes permit revisitation and loop formation to maintain solution feasibility.

Unlike the profit-to-time greedy construction in \citet{Gong2025Deep} for PDRA-Basic, the temporal constraints in PDRA-TW necessitate time-window-aware selection criteria and feasibility verification at both construction and improvement phases. This variant-specific algorithmic adaptation exemplifies the domain knowledge requirements that traditional methods impose, requirements that our UM circumvents through learned representations. While the framework presented below could be adapted to other variants through modified feasibility checks (e.g., removing time-window constraints for PDRA-Basic, incorporating multi-depot assignment logic for PDRA-MD), we focus exposition on PDRA-TW to avoid redundancy while illustrating the fundamental challenges that motivate the learning-based approach.

\subsection*{Phase 1: Time-window-aware greedy construction}

To generate an initial feasible solution, we develop a construction heuristic that balances information value collection with time-window urgency. For each drone $k$, starting from the depot $o$, the algorithm iteratively selects the next artificial node $p \in \mathcal{P} \setminus \mathcal{V}$ based on a composite scoring function that incorporates both value-efficiency and temporal urgency:

\begin{equation}
\label{D1}
\text{score}(p) = \alpha \cdot \frac{c_p}{\tau_{up}} + \beta \cdot \frac{1}{l_p - (d_t + \tau_{up}) + \epsilon},
\tag{D1}
\end{equation}

\noindent where $c_p$ denotes the information value of artificial node $p$; $\tau_{up}$ represents the shortest path time from current node $u$ to artificial node $p$ in the transformed network $\bar{G}$; $d_t$ is the accumulated flight time of the current drone; $l_p$ is the latest allowable assessment time at node $p$; $\epsilon$ is a small constant (e.g., $\epsilon = 0.01$) to ensure numerical stability; and $\alpha$ and $\beta$ are weighting parameters balancing value-efficiency and temporal urgency, set to $\alpha = 0.7$ and $\beta = 0.3$ based on preliminary experiments. The first term $c_p / \tau_{up}$ represents the value-to-time ratio, favoring nodes that yield high information value per unit travel time. The second term captures temporal urgency, prioritizing nodes whose time windows are approaching expiration. This composite score enables the heuristic to adaptively balance between maximizing collected value and maintaining temporal feasibility. Additionally, the node selection is subject to the following feasibility constraints. First, the candidate node must be reachable from the current node through the transformed network $\bar{G}$. Second, the time window must remain valid upon arrival, that is, $d_t + \tau_{up} \le l_p$. Third, the remaining mission time must be sufficient to complete the assessment at $p$ and return to the depot, requiring $d_t + \tau_{up} + \tau_{po} \le \min{Q, p_{\max}}$. Finally, each artificial node can be visited only once, ensuring exclusivity across drones.

Once $p$ is selected, the drone traverses from $u$ to $p$, completes the assessment, and reaches the exit original node $v$. The algorithm then updates the current node to $u \leftarrow v$, the accumulated time to $d_t \leftarrow d_t + \tau_{up}$, and adds $p$ to the visited set $\mathcal{V}$. This process continues until no more time-window-feasible nodes remain, after which the drone returns to the depot and the next drone begins route construction. The construction phase terminates when either all $K$ drones have been assigned routes or all artificial nodes with positive information values have been visited.

\subsection*{Phase 2: Time-window-aware local search}

Following the construction of an initial solution, an iterative refinement procedure explores neighboring configurations to enhance the objective value while preserving time-window feasibility. In the transformed network representation, each artificial node $p \in \mathcal{P}$ is embedded within a triplet structure $(u_p, p, v_p)$, where $u_p$ and $v_p$ represent the preceding and succeeding original nodes along the route. The improvement phase employs three search operators designed for the triplet-based network structure. Relocate operator modifies the traversal direction of a triplet by reversing its orientation from $(u_p, p, v_p)$ to $(v_p, p, u_p)$ within a single route, thereby investigating alternative access paths to the assessed road segment. Remove-Insert operator eliminates an existing triplet $(u_p, p, v_p)$ and incorporates an unvisited triplet $(u_q, q, v_q)$ with $q \in \mathcal{P} \setminus \mathcal{V}$ into the route at an alternative location, enabling the substitution of lower-value assessments with potentially higher-value alternatives. Exchange operator performs a bilateral swap of triplets $(u_p, p, v_p)$ and $(u_q, q, v_q)$ across distinct drone routes, reallocating assessment responsibilities to enhance the global solution quality.

Each operator application requires validation of two conditions: (i) temporal feasibility, ensuring that revised arrival times $d'_t(i)$ satisfy $d'_t(i) \leq l_i$ for all nodes in modified route segments, and (ii) capacity compliance, verifying that updated routes adhere to $d_t + \tau_{ij} + \tau_{jo} \leq \min\{Q, p_{\max}\}$. The search procedure adopts an accept-first policy wherein operators are examined in sequence, and the initial improvement-yielding feasible neighbor is immediately adopted. Exploration continues for $\text{max\_iterations} = 1{,}000$ cycles. Algorithm~\ref{algorithmTW} presents the complete procedure for PDRA-TW.

\begin{algorithm}[!ht]
\caption{Two-Phase Heuristic for PDRA-TW with Time Windows}
\label{algorithmTW}
\begin{algorithmic}[1]
\STATE \textbf{Input:} Transformed graph $\bar{G}$, travel times $\tau_{ij}$, node values $c_p$, depot $o$, deadlines $l_p$, fleet size $K$, capacity bounds $Q, p_{\max}$, iteration limit $\text{max\_iterations}$
\vspace{0.4em}

\STATE \textbf{Phase 1: Greedy construction with temporal awareness}
\STATE Initialize: $\mathcal{V} \leftarrow \emptyset$, $\Pi \leftarrow \emptyset$
\FOR{each drone $k \in \{1, \ldots, K\}$}
    \STATE Current position $u \leftarrow o$, elapsed time $d_t \leftarrow 0$
    \WHILE{feasible candidate nodes remain}
        \STATE Choose $p^* = \arg\max_{p \in \mathcal{P} \setminus \mathcal{V}} \text{score}(p)$ via Equation~\eqref{D1}
        \STATE Append path $u \rightsquigarrow p^*$ to partial route $\pi^k$
        \STATE Update: $d_t \leftarrow d_t + \tau_{up^*}$, $\mathcal{V} \leftarrow \mathcal{V} \cup \{p^*\}$, $u \leftarrow v^*$
    \ENDWHILE
    \STATE Complete route with return to $o$, add $\pi^k$ to $\Pi$
\ENDFOR

\vspace{0.4em}
\STATE \textbf{Phase 2: Iterative refinement with temporal constraints}
\STATE Best solution: $\Pi^* \leftarrow \Pi$, optimal value $R^* \leftarrow R(\Pi)$
\FOR{iteration $\text{iter} \in \{1, \ldots, \text{max\_iterations}\}$}
    \FOR{each operator in $\{\text{Relocate}, \text{Remove-Insert}, \text{Exchange}\}$}
        \STATE Construct candidate $\Pi'$ through operator application
        \IF{$\Pi'$ satisfies constraints \textbf{and} $R(\Pi') > R(\Pi)$}
            \STATE Accept improvement: $\Pi \leftarrow \Pi'$
            \IF{$R(\Pi') > R^*$}
                \STATE Record new best: $\Pi^* \leftarrow \Pi'$, $R^* \leftarrow R(\Pi')$
            \ENDIF
            \STATE \textbf{break} \COMMENT{Accept first improvement}
        \ENDIF
    \ENDFOR
\ENDFOR
\STATE \textbf{Return:} Optimized solution $\Pi^*$ with value $R^*$
\end{algorithmic}
\end{algorithm}

\end{document}